\documentclass[twoside,11pt]{article}

\usepackage{jmlr2e}

\usepackage{amsmath}
\usepackage{graphicx}  
\usepackage{float}
\usepackage{xcolor}
\usepackage{colortbl}
\usepackage{subcaption}
\usepackage[font={small,it}]{caption}
\usepackage{booktabs}
\usepackage{algorithm}
\usepackage[noend]{algpseudocode}

\newcommand{\toponecell}{\cellcolor{red!50!white}}
\newcommand{\topthreecell}{\cellcolor{red!20!white}}

\newcommand{\qedwhite}{\hfill \ensuremath{\Box}}
\newcommand{\cb}{\color{black}}
\newcommand{\cred}{\color{black}}

\newtheorem{assumption}[theorem]{Assumption}

\begin{document}

\title{Diminishing Batch Normalization}

\author{\name Yintai Ma \email yintaima2020@u.northwestern.edu \\
       \addr Department of Industrial Engineering \\
       Northwestern University\\
       Evanston, IL 60208, USA
       \AND
       \name Diego Klabjan \email d-klabjan@northwestern.edu \\
       \addr Department of Industrial Engineering \& Management Science\\
       Northwestern University\\
       Evanston, IL 60208, USA}
       
\editor{TBD.}

\maketitle

\begin{abstract}
In this paper, we propose a generalization of the Batch Normalization (BN) algorithm, diminishing batch normalization (DBN), where we update the BN parameters in a diminishing moving average way. BN is very effective in accelerating the convergence of a neural network training phase that it has become a common practice. 
Our proposed DBN algorithm remains the overall structure of the original BN algorithm while introduces a weighted averaging update to some trainable parameters. 
We provide an analysis of the convergence of the DBN algorithm that converges to a stationary point with respect to trainable parameters. Our analysis can be easily generalized for original BN algorithm by setting some parameters to constant. To the best knowledge of authors, this analysis is the first of its kind for convergence with Batch Normalization introduced. We analyze a two-layer model with arbitrary activation function. 
The convergence analysis applies to any activation function that satisfies our common assumptions.
In the numerical experiments, we test the proposed algorithm on complex modern CNN models with stochastic gradients and ReLU activation. We observe that DBN outperforms the original BN algorithm on MNIST, NI and CIFAR-10 datasets with reasonable complex FNN and CNN models.
\end{abstract}


\section{Introduction}

	Deep neural networks (DNN) have shown unprecedented success in various  applications such as object detection. However, it still takes a long time to train a DNN until it converges. { \cite{SergeyIoffe2015} identified a critical problem involved in training deep networks, internal covariate shift, and then proposed batch normalization (BN) to decrease this phenomenon. BN addresses this problem by normalizing the distribution of every hidden layer's input. In order to do so, it calculates the pre-activation mean and standard deviation using mini-batch statistics at each iteration of training and uses these estimates to normalize the input to the next layer. The output of a layer is normalized by using the batch statistics, and two new trainable parameters per neuron are introduced that capture the inverse operation. It is now a standard practice \cite{Bottou2016,He2015a}. 
		While this approach leads to a significant performance jump, to the best of our knowledge, there is no known theoretical guarantee for the convergence of an algorithm with BN. The difficulty of analyzing the convergence of the BN algorithm comes from the fact that not all of the BN parameters are updated by gradients. Thus, it invalidates most of the classical studies of convergence for gradient methods.}
	
	In this paper, we propose a generalization of the BN algorithm, diminishing batch normalization (DBN), where we update the BN parameters in a diminishing moving average way. It essentially means that the BN layer adjusts its output according to all past mini-batches instead of only the current one. 
	It helps to reduce the problem of the original BN that the output of a BN layer on a particular training pattern depends on the other patterns in the current mini-batch, which is pointed out by \cite{Bottou2016}.
	By setting the layer parameter we introduce into DBN to a specific value, we recover the original BN algorithm.
	
	We give a convergence analysis of the algorithm with a two-layer batch-normalized neural network and diminishing stepsizes.
	We assume two layers (the generalization to multiple layers can be made by using the same approach but substantially complicating the notation) and an arbitrary loss function. The convergence analysis applies to any activation function that follows our common assumption. The main result shows that under diminishing stepsizes on gradient updates and updates on mini-batch statistics, and standard Lipschitz conditions on loss functions DBN converges to a stationary point. As already pointed out the primary challenge is the fact that some trainable parameters are updated by gradient while others are updated by a minor recalculation. 
	
	\textbf{Contributions.} \quad The main contribution of this paper is in providing a general convergence guarantee for DBN. Specifically, we make the following contributions.
	\begin{itemize}
		\item In Section \ref{sec:nonconvex}, we show conditions for the stepsizes and diminishing weights to ensure the convergence of BN parameters. The proof is provided in the appendix.
		\item We show that the algorithm converges to a stationary point under a general nonconvex objective function. To the best of our knowledge, this is the first convergence analysis that specifically considers transformations with BN layers. 
	\end{itemize}
	
	This paper is organized as follows. In Section \ref{sec:lr}, we review the related works and the development of the BN algorithm. We formally state our model and algorithm in Section \ref{sec:model_algorithm}. We present our main results in Sections \ref{sec:nonconvex}. In Section \ref{sec:computation}, we numerically show that the DBN algorithm outperforms the original BN algorithm.  Proofs for main steps are collected in the Appendix.

	\section{Literature Review}
		\label{sec:lr}
		

		Before the introduction of BN, it has long been known in the deep learning community that
		input whitening and decorrelation help to speed up
		the training process. In fact, \cite{Orr2003} show that preprocessing the data by subtracting the mean, normalizing
		the variance, and decorrelating the input has various beneficial effects for back-propagation.
		\cite{Krizhevsky2012} propose a method called local response normalization
		which is inspired by computational neuroscience and acts as a form of lateral inhibition,
		i.e., the capacity of an excited neuron to reduce the activity of its neighbors. 
		\cite{Gulcehre2016} propose a standardization layer that bears significant resemblance
		to batch normalization, except that the two methods are motivated by
		very different goals and perform different tasks.
		
		
		
		Inspired by BN, several new works are taking BN as a basis for further improvements.
		Layer normalization \cite{Ba} is much like the BN except that it uses all of the summed inputs to compute the mean and variance instead of the mini-batch statistics. Besides, unlike BN, layer normalization performs precisely the same computation at training and test times.
		Normalization propagation that \cite{Arpit2016} uses data-independent estimations for the mean and standard deviation in every layer to reduce the internal covariate shift and make the estimation more accurate for the validation phase.
		Weight normalization also removes the dependencies between the examples in a minibatch so that it can be applied to recurrent models, reinforcement learning or generative models \cite{Salimans}. 
		\cite{Cooijmans} propose a new way to apply batch normalization to RNN and LSTM models.
 Recently, there are works on the insights and analysis of Batch Normalization. \cite{Johan2018} demonstrate how BN can help to correct training for ill-behaved normalized networks. \cite{santurkar2018does} claim that the key factor of BN is that it makes the optimization landscape much smoother and allows for faster training. However, these works do not cover a convergence analysis for BN and hence do not overlap with this work.

		Given all these flavors, the original BN method is the most popular technique and for this reason our choice of the analysis. To the best of our knowledge, we are not aware of any prior analysis of BN.

		BN has the gradient and non-gradient updates. Thus, nonconvex convergence results do not immediately transfer. Our analysis explicitly considers the workings of BN. However, nonconvex convergence proofs are relevant since some small portions of our analysis rely on known proofs and approaches.
		
		Neural nets are not convex, even if the loss function is convex.  
		For classical convergence results with a nonconvex objective function and diminishing learning rate, we refer to survey papers \cite{Bertsekas2010,Bertsekas,Bottou2016}. 
		\cite{Bertsekas} provide a convergence result with the deterministic gradient with errors. 
		\cite{Bottou2016} provide a convergence result with the stochastic gradient. 
		The classic analyses showing the norm of gradients of the objective function going to zero date back to \cite{Grippo1994,Polyak1973,Polyak1987}.
		For strongly convex objective functions with a diminishing learning rate, we learn the classic convergence results from 
		\cite{Bottou2016}.

				\section{Model and Algorithm}
				\label{sec:model_algorithm}
				The optimization problem for a network is an objective function consisting of a large number of component functions, that reads:
				\begin{equation}
				\label{dfn:our_model}
				\begin{aligned}
				\min \ & \bar{f}(\theta, \lambda) = \sum_{i=1}^{N} f_i(X_i:\theta,\lambda), 
				\end{aligned}
				\end{equation}
				where $ f_i : \mathbb{R}^{n_1} \times \mathbb{R}^{n_2} \rightarrow \mathbb{R}, i =1 ,...,N$, are real-valued functions for any data record $ X_i $.
				Index $ i $ associates with data record $ X_i $ and target response $ y_i $ 
				(hidden behind the dependency of $ f $ on $ i $)
				in the training set. 
				Parameters $ \theta $ include the common parameters updated by gradients directly associated with the loss function, while BN parameters $ \lambda $ are introduced by the BN algorithm and not updated by gradient methods but by mini-batch statistics.
				We define that the derivative of $ f_i $ is always taken with respect to $ \theta $:
				\begin{equation}
				\nabla  f_i(X_i: \theta,\lambda) := \nabla_\theta  f_i(X_i: \theta, \lambda).
				\end{equation}
								
				The deep network we analyze has 2 fully-connected layers with $ D $ neurons each. The techniques presented can be extended to more layers with additional notation. Each hidden layer computes $ y=a(Wu) $ with nonlinear activation function $ a(\cdot) $ and $ u $ is the input vector of the layer. We do not need to include an intercept term since the BN algorithm automatically adjusts for it. BN is applied to the output of the first hidden layer. 	
				
				\begin{figure}[H]
					\centering
					\includegraphics[width=0.7\linewidth]{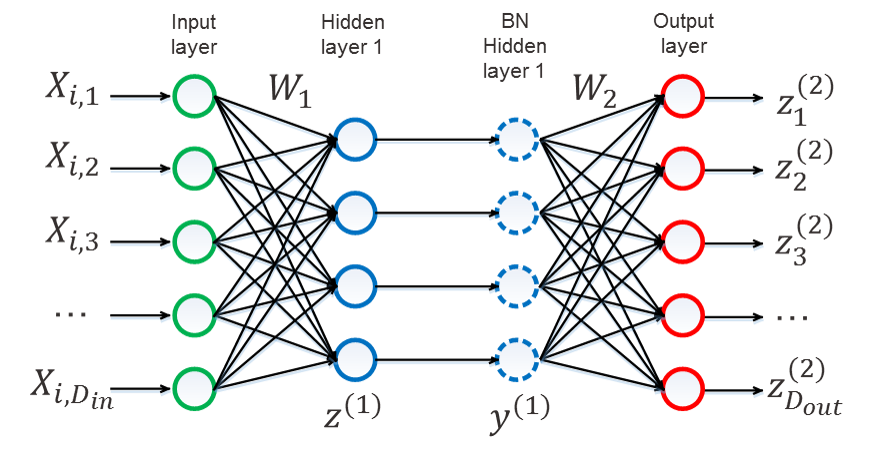}
					\caption{The structure of our batch-normalized network model in the analysis.}
					\label{fig:NetworkStructure}
				\end{figure}

				We next describe the computation in each layer to show how we obtain the output of the network. 
				The notations introduced here is used in the analysis. 
				Figure \ref{fig:NetworkStructure} shows the full structure of the network. 
				The input data is vector $ X $, which is one of $ \{X_i\}_{i=1}^N $. 
				Vector $ \lambda = \left(  
				(\mu_j)_{j=1}^{D} , (\sigma_j)_{j=1}^{D}
				\right)
				$
				is the set of all BN parameters and 
				vector $ \theta = \left(
				W_1, W_2, 
				(\beta_j^{(1)})_{j=1}^{D}, (\gamma_j^{(1)})_{j=1}^{D}
				\right)
				$
				is the set of all trainable parameters which are updated by gradients.
				
				{ 
					Matrices $ W_1, W_2 $ are the actual model parameters and $ \beta, \gamma $ are introduced by BN.
					The value of $ j^{th} $ neuron of the first hidden layer is
					\begin{equation}
					z_j^{(1)} (X:\theta) = a(W_{1,j,\cdot} X),
					\end{equation}
					where $ W_{1,j,\cdot} $ denotes the weights of the linear transformations for the $ j^{th} $ neuron. 
					
					The $ j^{th} $ entry of batch-normalized output of the first layer is
					\begin{align*}
					y_j^{(1)}(X : \theta,\lambda)  &  = \gamma_j^{(1)} \left(\frac{z_j^{(1)}(X:\theta)-\mu_j}{\sigma_j + \epsilon_B}\right) +\beta_j^{(1)},
					\end{align*}
					where $ \beta_j^{(1)} $ and $ \gamma_j^{(1)} $ are trainable parameters updated by gradient and $ \mu_j $ and $ \sigma_j $ are batch normalization parameters for $ z_j^{(1)} $. Trainable parameter $ \mu_j $ is the mini-batch mean of $ z_j^{(1)} $ and trainable parameter $ \sigma_j $ is the mini-batch sample deviation of $ z_j^{(1)} $. Constant $ \epsilon_B $ keeps the denominator from zero. 
					The output of $ j^{th} $ entry of the output layer is:
					\begin{equation}
						z_j^{(2)} (X:\theta) = 
						a\left(W_{2,j,\cdot} \left[ \gamma_j^{(1)} \left(\frac{z_j^{(1)}(X:\theta)-\mu_j}{\sigma_j + \epsilon_B}\right) +\beta_j^{(1)}\right]\right)
					\end{equation}
					The objective function for the $ i^{th} $ sample is
					\begin{equation}
					\label{dfn:cost_function_general}
					f_i(X_i: \theta, \lambda) = 
					l_i\left( \left( z_j^{(2)}(X_i: \theta,\lambda) \right)_j \right),
					\end{equation} 
					where $ l_i(\cdot) $ is the loss function associated with 
					the target response $ y_i $. 
					For sample $i$, 
					we have the following complete expression for the objective function:
					\begin{equation}
					\label{dfn:cost_function_exact}
					f_i(X_i: \theta, \lambda) = 
					l_i \left(
					a( 
					\sum_{j=1}^{D}W_{2,k,j} \left[ \gamma_j^{(1)} \frac{ a(W_{1, j,\cdot} X_i -\mu_j) }{\sigma_j+ \epsilon_B} + \beta_j^{(1)} \right] )_k   \right).
					\end{equation}
					Function $ f_i(X_i:\theta, \lambda) $ is nonconvex with respect to $ \theta $ and $ \lambda $.
				}
				
				\subsection{Algorithm}
				
				Algorithm \ref{alg_BNIGA} shows the algorithm studied herein. There are two deviations from the standard BN algorithm, one of them actually being a generalization. 
				We use the full gradient instead of the more popular stochastic gradient (SG) method. 
				It essentially means that each batch contains the entire training set instead of a randomly chosen subset of the training set. 
				An analysis of SG is potential future research.
				Although the primary motivation for full gradient update is to reduce the burdensome in showing the convergence, the full gradient method is similar to SG in the sense that both of them go through the entire training set, while full gradient goes through it deterministically and the SG goes through it in expectation. Therefore, it is reasonable to speculate that the SG method has similar convergence property as the full algorithm studied herein. 
				
				
				\begin{algorithm}[H]
					\caption{DBN: Diminishing Batch-Normalized Network Update Algorithm}\label{alg_BNIGA}  
					\begin{algorithmic}[1]
						\State Initialize $ \theta \in \mathbb{R}^{n_1}$ and $ \lambda \in \mathbb{R}^{n_2} $
						\For {iteration $ m $=1,2,... } 
						\State $ \theta^{(m+1)} := \theta^{(m)} - \eta^{(m)} \sum_{i=1}^{N} \nabla  f_i(X_i: \theta^{(m)},\lambda^{(m)})$
						\For {$ j $=1,...,$ D_1 $}
						\State $ \mu_j^{(m+1)} := \frac{1}{N} \sum_{i=1}^{N}  z_j^{(1)}(X_{i}:\theta^{(m+1)})  $
						\State 
						$ \sigma_j^{(m+1)} := \sqrt{\frac{1}{N} \sum_{i=1}^{N} \left( z_j^{(1)}(X_{i}:\theta^{(m+1)}) - \mu_j^{(m+1)} \right)^2 } $
						\EndFor
						\State $ \lambda^{(m+1)} := \alpha^{(m+1)} \left( (\mu_j^{(m+1)})_{j=1}^{D_1} , (\sigma_j^{(m+1)})_{j=1}^{D_1} \right) + (1-\alpha^{(m+1)})\lambda^{(m)}   $  
						\EndFor
					\end{algorithmic}
				\end{algorithm}
				
				The second difference is that we update the BN parameters $ (\theta, \lambda) $ by their moving averages 
				with respect to diminishing $ \alpha^{(m)}$. The original BN algorithm can be recovered by setting
				$ \alpha^{(m)} = 1 $ for every $m$. 
				After introducing diminishing $ \alpha^{(m)} $, $ \lambda^{(m)} $ and hence the output of the BN layer is determined by the history of all past 
				data records, instead of those solely in the last batch. Thus, the output of the BN layer becomes more general that better reflects the distribution of the entire dataset. 
				We use two strategies to decide the values of $ \alpha^{(m)} $. One is to use a constant smaller than 1 for all $ m $, and the other one is to decay the $ \alpha^{(m)} $ gradually, such as $ \alpha^{(m)}=1/m$.

				In our numerical experiment, we show that Algorithm \ref{alg_BNIGA} outperforms the original BN algorithm, where both are based on SG and non-linear activation functions with many layers FNN and CNN models.

				\section{General Case}
				\label{sec:nonconvex}
				
				The main purpose of our work is to show that Algorithm \ref{alg_BNIGA} converges.
				In the general case, we focus on the nonconvex objective function.
				
				\subsection{Assumptions}
				
				Here are the assumptions we used for the convergence analysis.

				\begin{assumption}
					\textbf{\text{(Lipschitz continuity on $ \theta $ and $ \lambda $).}} 
					\label{ass:lipschitz_theta}
					For every $ i $ we have
					\begin{equation}
					\lVert \nabla f_i(X:\tilde{\theta}, \lambda) - \nabla f_i(X:\hat{\theta}, \lambda) \rVert_2  \leq \bar{L} \lVert \tilde{\theta} - \hat{\theta} \rVert_2, \forall \tilde{\theta},\hat{\theta},\lambda,X.
					\label{ass:L_1_0}
					\end{equation}
					\begin{equation}
					\begin{aligned}
					\lVert \nabla_{W_{1,j,\cdot}} f_i(X:\tilde{\theta}, \lambda) - \nabla_{W_{1,j,\cdot}} f_i(X:\hat{\theta}, \lambda) \rVert_2  \\
					\leq \bar{L} \lVert \tilde{W}_{1,j,\cdot} - \hat{W}_{1,j,\cdot} \rVert_2,
					\forall \lambda, \tilde{\theta}, \hat{\theta},X, j \in \{1,...,D_1\}.
					\end{aligned}
					\label{ass:L_1}
					\end{equation}
					\begin{equation}
					\begin{aligned}
					\lVert \nabla f_i(X:\theta, \tilde{\lambda}) - \nabla f_i(X:\theta, \hat{\lambda}) \rVert_2  \leq \bar{L} \lVert \tilde{\lambda} - \hat{\lambda} \rVert_2,  \\
					\forall \theta, \tilde{\lambda}, \hat{\lambda},X, j \in \{1,...,D_1\}.
					\end{aligned}
					\label{ass:L_4}
					\end{equation}
				\end{assumption}
				Noted that the Lipschitz constants associated with each of the above inequalities are not necessarily the same. Here $ \bar{L} $ is an upper bound for these Lipschitz constants for simplicity.

				\begin{assumption}
					\textbf{\text{(bounded parameters).}} \label{ass:bounded_para}
					There exists a constant $ M $ such that weights $ W^{(m)} $ and parameters $ \lambda^{(m)} $ are bounded element-wise by this constant $ M $ in every iteration $ m $, 
					\[ \lVert W_{1}^{(m)} \rVert \preceq M \text{ and } \lVert W_2^{(m)} \rVert \preceq M \text{ and }  \lVert \lambda^{(m)} \rVert \preceq M .\]
				\end{assumption}
				
				
				
				

				%
				%
				
				\begin{assumption}
					\textbf{\text{(diminishing update on $ \theta $).}} \label{ass:diminish_theta_update}
					The stepsizes of $ \theta $ update satisfy
					\begin{equation}
					\sum_{m=1}^{\infty} \eta^{(m)} = \infty \ and \ \sum_{m=1}^{\infty} (\eta^{(m)})^2 < \infty.
					\label{res:diminish_theta_update}
					\end{equation}
				\end{assumption}
				
				This is a common assumption for diminishing stepsizes in optimization problems.
				
				\begin{assumption}
					\textbf{\text{(Lipschitz continuity of $ l_i(\cdot) $).}} \label{ass:l_lipschitz}
					Assume the loss functions  $ l_i(\cdot) $ for every $ i $ is continuously differentiable. It implies that 
					there exists $ \hat{M} $ such that 
					\[ \lVert l_i(x) - l_i(y)\rVert \leq \hat{M} \lVert x - y \rVert, \forall x,y. \]
				\end{assumption}
				
				\begin{assumption}
					\textbf{\text{(existence of a stationary point).}} \label{ass:stationary_point}
					There exists a stationary point $ (\theta^*, \lambda^*) $ such that $ \lVert \nabla \bar{f}(\theta^*, \lambda^*) \rVert = 0. $
				\end{assumption}
				
				We note that all these are standard assumptions in convergence proofs. We also stress that Assumption \ref{ass:l_lipschitz} does not directly imply \ref{ass:lipschitz_theta}. 
				Assumptions \ref{ass:lipschitz_theta},
 				 \ref{ass:l_lipschitz} and \ref{ass:stationary_point} hold for many standard loss functions such as softmax and MSE.
 				
				
				{
					\begin{assumption}
						\textbf{\text{(Lipschitz at activation function).}} \label{ass:bounding_activation}
						The activation function $ a(\cdot) $ is Lipschitz with constant $ k $:
						\begin{equation}
						\lvert a(x) \rvert \leq k \lVert x \rVert
						\end{equation}		
					\end{assumption}
					Since for all activation function there is $ a(0)=0$, the condition is equivalent to $ 			\lvert a(x) - a(0) \rvert \leq k \lVert x - 0 \rVert $.
					We note that this assumption works for many popular choices of activation functions, such as ReLU and LeakyReLu.
				}

				\subsection{Convergence Analysis}

				We first have the following lemma specifying sufficient conditions for $ \lambda $ to converge. 
				Proofs for main steps are given in the Appendix.
				
				\begin{theorem}
					\label{thm:lambda_converge}
					Under Assumptions
					\ref{ass:lipschitz_theta}, \ref{ass:bounded_para}, \ref{ass:diminish_theta_update} and \ref{ass:bounding_activation},
					if $ \{ \alpha^{(m)}\} $ satisfies 
					\[ \sum_{m=1}^{\infty} \alpha^{(m)} <\infty  \text{  and  }
					\sum_{m=1}^{\infty} \sum_{n=1}^{m} \alpha^{(m)} \eta^{(n)} <\infty,
					\]
					then sequence $ \{ \lambda^{(m)} \} $ converges to $ \bar{\lambda} $.
					\label{thm:2_plus}
				\end{theorem}
				
				We show in Theorem \ref{thm:double_limits} that this 
				$ \bar{\lambda}  $ converges to $ \lambda^* $, where the loss function reaches zero gradients, i.e., $ (\theta^*, \lambda^*) $ is a stationary point.
				We give a discussion of the above conditions for $ \alpha^{(m)} $ and $ \eta^{(m)} $ at the end of this section.
				With the help of Theorem \ref{thm:lambda_converge}, we can show the following convergence result.
				
				\begin{lemma}
					\label{thm:nonconvex_diminishing2}
					Under 
					Assumptions \ref{ass:l_lipschitz}, \ref{ass:stationary_point}
					and 
					the assumptions of Theorem \ref{thm:lambda_converge}, 
					when
					\begin{equation}
					\sum_{m=1}^{\infty} \sum_{i=m}^{\infty} \sum_{n=1}^{i} \alpha^{(i)} \eta^{(n)} <\infty \quad \text{and} \quad  
					\sum_{m=1}^{\infty} \sum_{n=m}^{\infty} \alpha^{(n)} <\infty,
					\label{condition:thm4.7}
					\end{equation}
					we have
					\begin{equation}
					\limsup\limits_{M\rightarrow \infty}
					\sum_{m=1}^{M} \eta^{(m)} \lVert \nabla \bar{f}(\theta^{(m)}, \bar{\lambda}) \rVert_2^2 < \infty.
					\label{res:nonconvex_res_1}
					\end{equation}
				\end{lemma}
				
				This result is similar to the classical convergence rate analysis for the non-convex objective function with diminishing stepsizes, which can be found in \cite{Bottou2016}.
				
				\begin{lemma}
					\label{thm:nonconvex_diminishing}
					Under the assumptions of Lemma \ref{thm:nonconvex_diminishing2}, we have
					\begin{equation}
					\liminf\limits_{m \rightarrow \infty} \lVert \nabla \bar{f}(\theta^{(m)}, \bar{\lambda}) \rVert^2_2 = 0.
					\label{res:nonconvex_convergence}
					\end{equation}
				\end{lemma}
				
				This theorem states that for the full gradient method with diminishing stepsizes the gradient norms cannot stay bounded away from zero. The following result characterizes more precisely the convergence property of Algorithm \ref{alg_BNIGA}.
				
				\begin{lemma}
					\label{lem:nonconvex_limit_gradient_zero}
					Under the assumptions stated in Lemma \ref{thm:nonconvex_diminishing2}, we have
					\begin{equation}
					\lim\limits_{m\rightarrow \infty} \lVert \nabla \bar{f}(\theta^{(m)}, \bar{\lambda}) \rVert_2^2 =0.
					\end{equation}
				\end{lemma}
				
				Our main result is listed next.

				\begin{theorem} \label{thm:double_limits}
					Under the assumptions stated in Lemma \ref{thm:nonconvex_diminishing2}, we have
					\begin{equation}
					\lim\limits_{m\rightarrow \infty} \lVert \nabla \bar{f}(\theta^{(m)}, \lambda^{(m)}) \rVert_2^2 =0.
					\end{equation}
				\end{theorem}
				
				It shows that the DBN algorithm converges to a stationary point where the norm of gradient is zero.
				We cannot show that $ \{\theta^{(m)}\} $'s converges (standard convergence proofs are also unable to show such a stronger statement). For this reason, Theorem \ref{thm:double_limits} does not immediately follow from Lemma \ref{lem:nonconvex_limit_gradient_zero} together with Theorem \ref{thm:lambda_converge}. 
				The statement of Theorem \ref{thm:double_limits} would easily follow from Lemma \ref{lem:nonconvex_limit_gradient_zero} if the convergence of $ \{\theta^{(m)}\} $ is established and the gradient being continuous.
				
				Considering the cases $ \eta^{(m)} = O(\frac{1}{m^k}) $ and $ \alpha^{(m)} = O(\frac{1}{m^h}) $. 
				We show in the appendix that the set of sufficient and necessary conditions to satisfy the  assumptions of Theorem \ref{thm:lambda_converge} are $ h>1 $ and $ k\geq1 $. 
				The set of sufficient and necessary conditions to satisfy the assumptions of Lemma \ref{thm:nonconvex_diminishing2} are $ h>2 $ and $ k\geq1 $. For example, we can pick 
				$ \eta^{(m)} = O(\frac{1}{m}) $ and $ \alpha^{(m)} = O(\frac{1}{m^{2.001}}) $ to achieve the above convergence result in Theorem \ref{thm:double_limits}.
				
				\section{Computational Experiments}
				\label{sec:computation}
				
				We conduct the computational experiments with Theano and Lasagne on a Linux server with a Nvidia Titan-X GPU.
				We use MNIST \cite{LeCun1998b}, CIFAR-10 \cite{Krizhevsky2009} and Network Intrusion (NI) \cite{kdd99} datasets to compare the performance between DBN and the original BN algorithm. For the MNIST dataset, we use a four-layer fully connected FNN ($ 784\times300\times300\times10 $) with the ReLU activation function and for the NI dataset, we use a four-layer fully connected FNN ($ 784\times50\times50\times10 $) with the ReLU activation function.
				For the CIFAR-10 dataset, we use a reasonably complex CNN network that has a structure of (Conv-Conv-MaxPool-Dropout-Conv-Conv-MaxPool-Dropout-FC-Dropout-FC),
				where all four convolution layers and the first fully connected layers are batch normalized. 
				We use the softmax loss function and $ l_2 $ regularization with for all three models. All the trainable parameters are randomly initialized before training. 
				{  For all 3 datasets, we use the standard epoch/minibatch setting with the minibatch size of $ 100 $, i.e., we do not compute the full gradient and the statistics are over the minibatch.}
				We use AdaGrad \cite{Duchi} to update the learning rates $ \eta^{(m)} $  for trainable  parameters, starting from $ 0.01 $. 
				
				We use two different strategies to decide the values of $ \alpha^{(m)} $ in DBN: constant values of $ \alpha^{(m)} $ and diminishing $ \alpha^{(m)} $ where $ \alpha^{(m)} = 1 / m $ and $ \alpha^{(m)} =  1/m^2 $.  We test the choices of constant $ \alpha^{(m)} \in \{1, 0.75, 0.5, 0.25, 0.1, 0.01, 0.001, 0\} $. 
				
				
					
				\begin{figure}[H]
					\begin{subfigure}{.33\textwidth}
						\centering
						\includegraphics[width=1\linewidth]{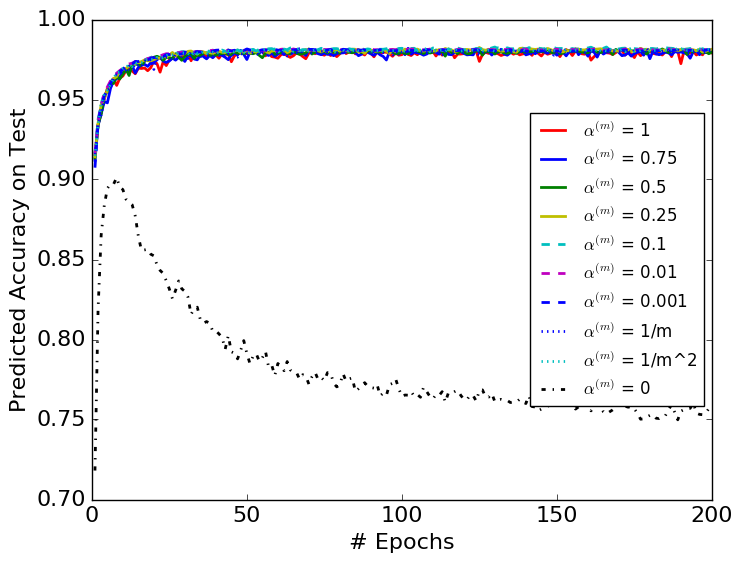}
						\caption{}
						\label{fig:sfig1_mnist_all_acc}
					\end{subfigure}%
					\begin{subfigure}{.33\textwidth}
						\centering
						\includegraphics[width=1\linewidth]{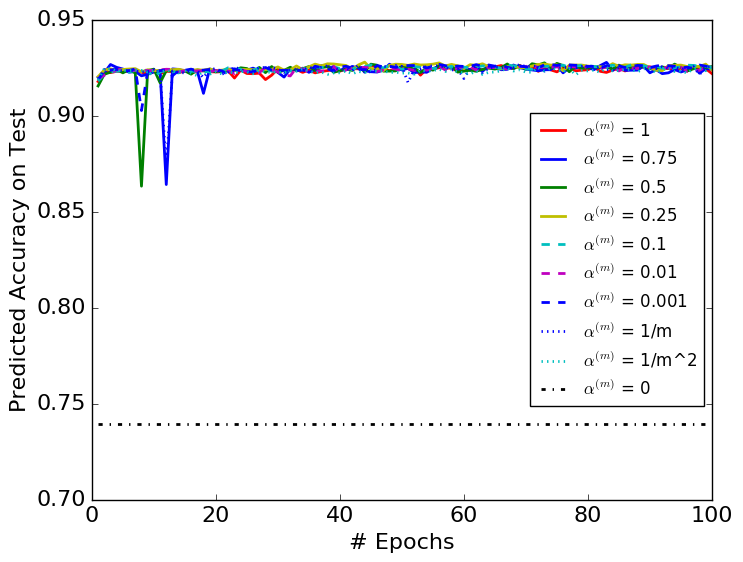}
						\caption{}
						\label{fig:sfig5_ni_all_acc}
					\end{subfigure}
					\begin{subfigure}{.33\textwidth}
						\centering
						\includegraphics[width=1\linewidth]{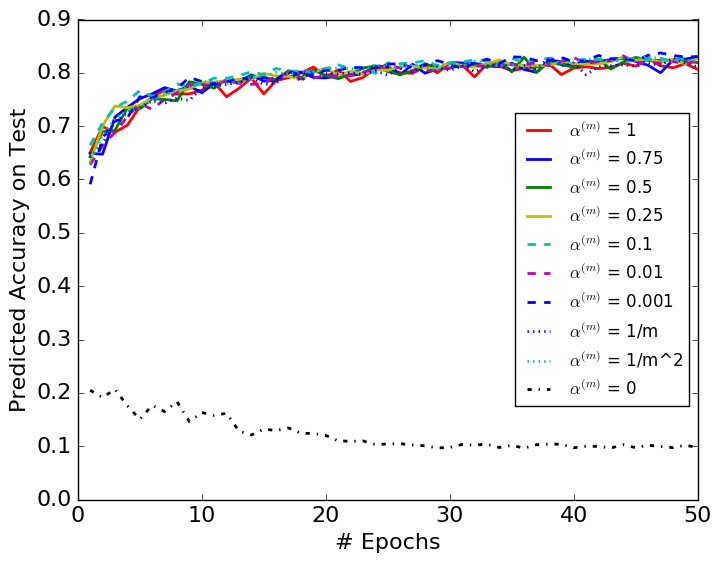}
						\caption{}
						\label{fig:sfig7_cifar_all_acc}
					\end{subfigure}%
					\caption{ Comparison of predicted accuracy on test datasets for different choices of $ \alpha^{(m)}$.
						From left to right are FNN on MNIST, FNN on NI and CNN on CIFAR-10. }
					\label{fig:all_acc}
				\end{figure}
				
				\begin{figure}[H]
					\begin{subfigure}{.33\textwidth}
						\centering
						\includegraphics[width=1\linewidth]{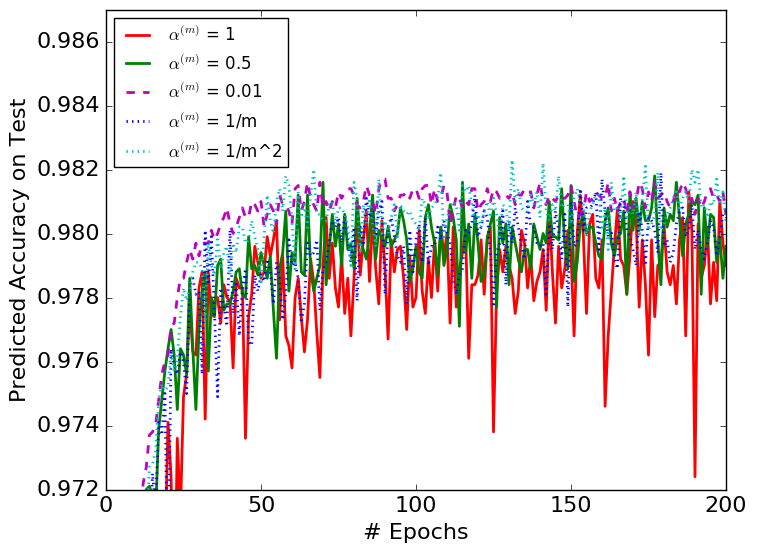}
						\caption{}
						\label{fig:sfig2_mnist_select_acc}
					\end{subfigure}
					\begin{subfigure}{.33\textwidth}
						\centering
						\includegraphics[width=1\linewidth]{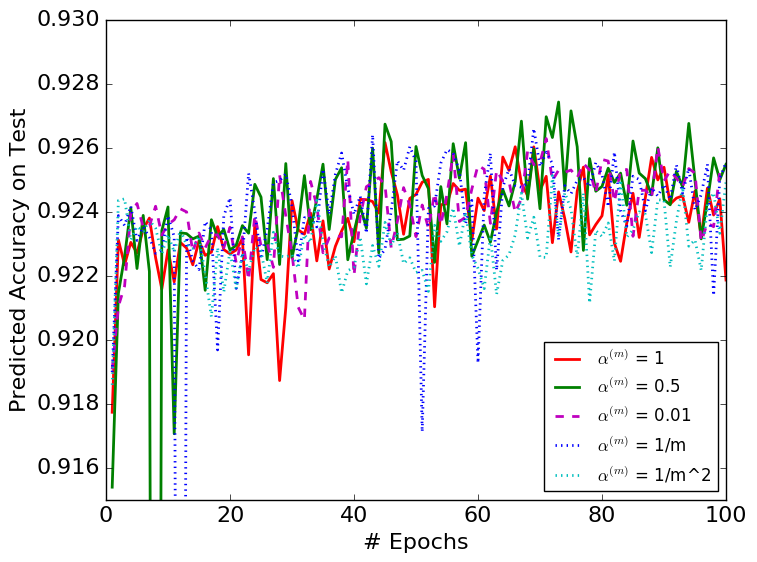}
						\caption{}
						\label{fig:sfig4_ni_select_acc}
					\end{subfigure}%
					\begin{subfigure}{.33\textwidth}
						\centering
						\includegraphics[width=1\linewidth]{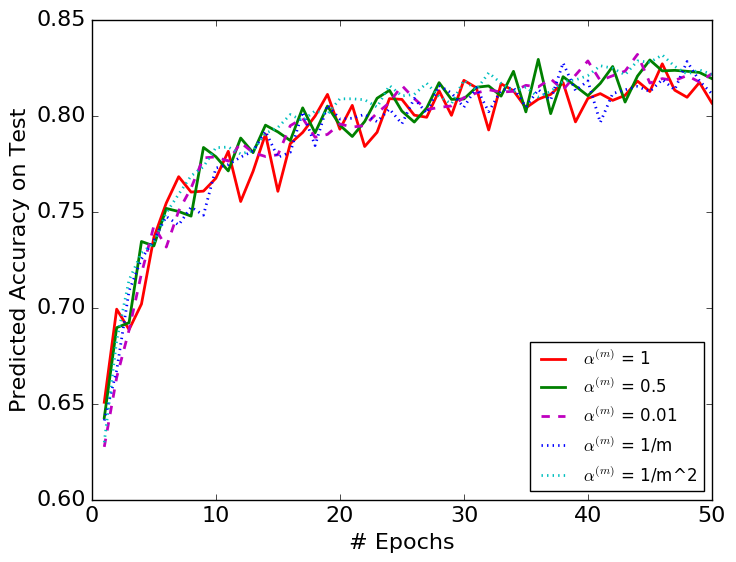}
						\caption{}
						\label{fig:sfig8_cifar_select_acc}
					\end{subfigure}
					\caption{ Comparison of predicted accuracy on test datasets for the most efficient choices of $ \alpha^{(m)}$.
						From left to right are  FNN on MNIST, FNN on NI and CNN on CIFAR-10. }
					\label{fig:select_acc}
				\end{figure}
				
				\begin{figure}[H]
					\begin{subfigure}{.32\textwidth}
						\centering
						\includegraphics[width=1\linewidth]{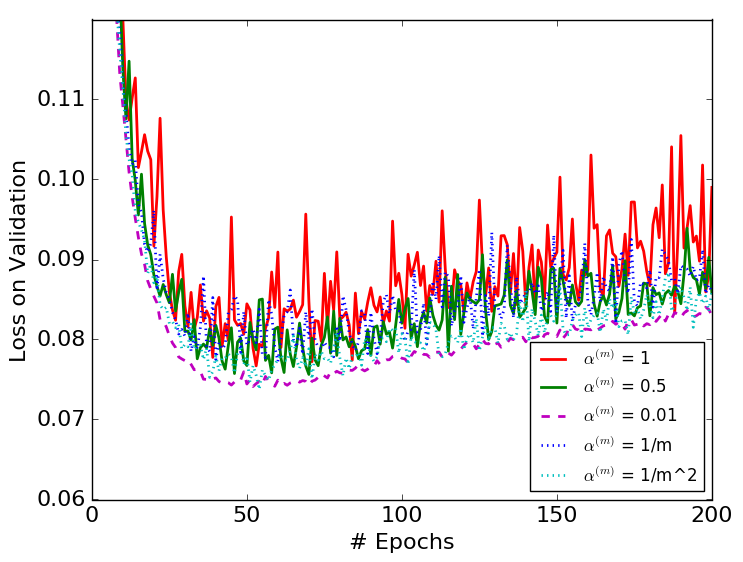}
						\caption{}
						\label{fig:sfig3_mnist_val_loss}
					\end{subfigure}
					\begin{subfigure}{.32\textwidth}
						\centering
						\includegraphics[width=1\linewidth]{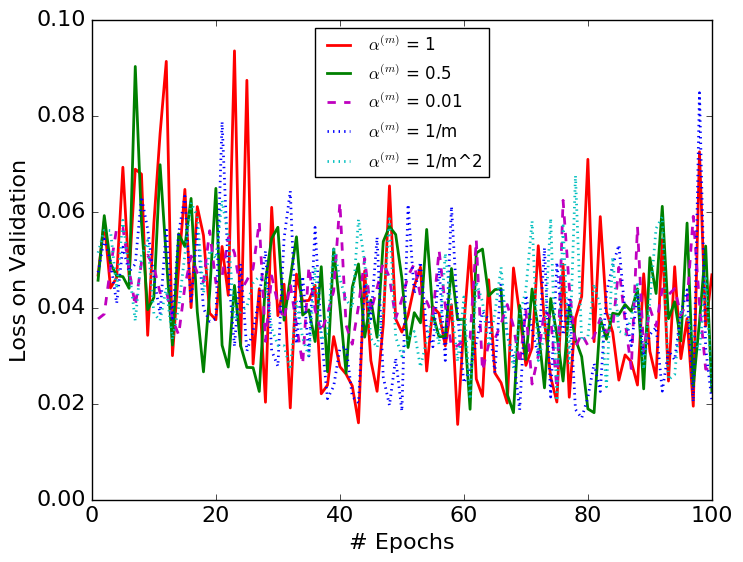}\\
						\caption{}
						\label{fig:sfig6_ni_val_loss}
					\end{subfigure}
					\begin{subfigure}{.32\textwidth}
						\centering
						\includegraphics[width=1\linewidth]{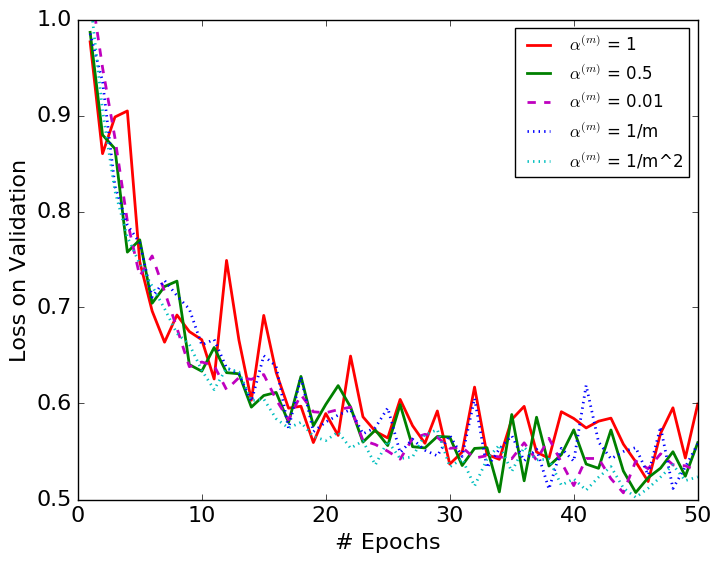}
						\caption{}
						\label{fig:sfig9_cifar_val_loss}
					\end{subfigure}
					\caption{ Comparison of the convergence of the loss function value on the validation set for different choices of $ \alpha^{(m)}$. 
						From left to right are FNN on MNIST, FNN on NI and CNN on CIFAR-10. }
					\label{fig:val_loss}
				\end{figure}
				
				We test all the choices of $ \alpha^{(m)} $ with the performances presented in Figure \ref{fig:all_acc}. 
				Figure \ref{fig:all_acc} shows that all the non-zero choices of $ \alpha^{(m)} $ converge properly. The algorithms converge without much difference even when $ \alpha^{(m)} $ in DBN is very small, e.g., $ 1/{m^2} $. However, if we select $ \alpha^{(m)}=0 $, the algorithm is erratic. 
				{ Besides, we observe that all the non-zero choices of $ \alpha^{(m)} $ converge at a similar rate. 
					The fact that DBN keeps the batch normalization layer stable with a very small $ \alpha^{(m)} $ suggests that the BN parameters do not have to be depended on the latest minibatch, i.e., the original BN.}
				
				We compare a selected set of the most efficient choices of $ \alpha^{(m)} $ in Figures \ref{fig:select_acc} and \ref{fig:val_loss}. They show that DBN with $ \alpha^{(m)}<1 $ is more stable than the original BN algorithm. The variances with respect to epochs of the DBN algorithm are smaller than those of the original BN algorithms in each figure.

				\begin{table}[htb]
					\scriptsize
					\centering
					\caption{Best results for different choices of $ \alpha^{(m)} $ on each dataset, showing the top three with a heat map.}
					\label{table:mnist_result}
						\begin{tabular}{lccc}
							\toprule
							& \multicolumn{3}{c}{ Test Error} \\
							\cmidrule{2-4}
							Model             & MNIST  & NI &  CIFAR-10 \\
							\midrule
							$\alpha^{(m)} = 1   $   & 2.70\%                         & 7.69\%                         &    17.31\%  \\
							$\alpha^{(m)} = 0.75$   &\topthreecell  1.91\%           & \topthreecell 7.37\%           &    17.03\%  \\
							$\alpha^{(m)} = 0.5 $   &\toponecell  \textbf{1.84\%}    & 7.46\%                         &    17.11\%  \\
							$\alpha^{(m)} = 0.25$   &\topthreecell  1.91\%           & \toponecell \textbf{7.24\%}    & \topthreecell    17.00\%  \\
							$\alpha^{(m)} = 0.1 $   &\topthreecell  1.90\%           &\topthreecell  7.36\%           &    17.10\%  \\
							$\alpha^{(m)} = 0.01$   & 1.94\%                         & 7.47\%                         &\topthreecell   16.82\%  \\
							$\alpha^{(m)} = 0.001$  & 1.95\%                         & 7.43\%                         &\toponecell   \textbf{16.28\%}  \\
							$\alpha^{(m)} = 1/m  $  & 2.10\%                         & 7.45\%                         &    17.26\%  \\
							$\alpha^{(m)} = 1/m^2$  & 2.00\%                         & 7.59\%                         &    17.23\%  \\
							$\alpha^{(m)} = 0    $ & 24.27\%                         & 26.09\%                        &     79.34\%   \\
							\bottomrule
						\end{tabular}
					\end{table}

					Table \ref{table:mnist_result} shows the best result obtained from each choice of $ \alpha^{(m)} $. 
					Most importantly, it suggests that the choices of $ \alpha^{(m)} = 1/m $ and $ 1/{m^2} $ perform better than the original BN algorithm. Besides, all the constant less-than-one choices of $ \alpha^{(m)} $ perform better than the original BN, showing the importance of considering the mini-batch history for the update of the BN parameters.  { The BN algorithm in each figure converges to similar error rates on test datasets with different choices of $ \alpha^{(m)} $ except for the $ \alpha^{(m)}=0 $ case. Among all the models we tested, $ \alpha^{(m)}=0.25 $ is the only one that performs top 3 for all three datasets, thus the most robust choice.}
					
					To summarize, our numerical experiments show that the DBN algorithm outperforms the original BN algorithm on the MNIST, NI and CIFAT-10 datasets with typical deep FNN and CNN models. 
					
					\textbf{Future Directions. \quad}
					On the analytical side, 
					we believe an extension to more than 2 layers is doable with significant augmentations of the notation.
					A stochastic gradient version is likely to be much more challenging to analyze.
					A second open question concerns analyzing the algorithm with a mini-batch setting. We believe it can be done by reusing most of the present analysis and changing some of the notation for wrapped layers.

\newpage

\appendix
\section*{Appendix A: Proofs.}

\subsection*{Preliminary Results}

{
\footnotesize
 \addtolength\abovedisplayskip{-0.35\baselineskip}%
 \addtolength\belowdisplayskip{-0.35\baselineskip}%

\begin{proposition}
	\label{pro:upper_bound_norm_gradient}
	{\cb There exists a constant M such that, for any $ \theta^{(m)} $ in iteration $ m $ and fixed $ \lambda $, we have
	\begin{equation}
	\lVert \nabla \bar{f}(\theta^{(m)}, {{\lambda}}) \rVert_2^2 \leq M.
	\end{equation} }
\end{proposition}
\textit{Proof.}
{ \cb 
By Assumption \ref{ass:stationary_point}, we know there exists $ (\theta^*, \lambda^*) $ such that $ \lVert \nabla \bar{f}(\theta^*, {\lambda^*})  \rVert_2 =  0 $. Then we have
\begin{equation}
\begin{aligned}
&\lVert \nabla \bar{f}(\theta^{(m)}, {\lambda})  \rVert_2  \\
= & \lVert \nabla \bar{f}(\theta^{(m)}, {\lambda})  \rVert_2 - \lVert \nabla \bar{f}(\theta^{(m)}, {\lambda^*})  \rVert_2 + \lVert \nabla \bar{f}(\theta^{(m)}, {\lambda^*})  \rVert_2 - \lVert \nabla \bar{f}(\theta^*, {\lambda^*})  \rVert_2  \\
\leq & 
\lVert \nabla \bar{f}(\theta^{(m)}, {\lambda})  - \nabla \bar{f}(\theta^{(m)}, {\lambda^*})  \rVert_2 
+ \lVert \nabla \bar{f}(\theta^{(m)}, {\lambda^*})   - \nabla \bar{f}(\theta^*, {\lambda^*})  \rVert_2   \\
 \leq  & \sum_{i=1}^{N} \lVert \nabla f_i(X_i:{\theta^{(m)}}, \lambda) - \nabla f_i(X_i:{\theta^{(m)}}, \lambda^*) \rVert_2 
	\\ & + \sum_{i=1}^{N} \lVert \nabla f_i(X_i:{\theta^{(m)}}, \lambda^*) - \nabla f_i(X_i:{\theta^*}, \lambda^*) \rVert_2 	\\
\leq &N\bar{L} (\lVert \lambda - \lambda^* \rVert_2 + \lVert \theta^{(m)} - \theta^* \rVert_2),
\end{aligned}
\end{equation}
where the last inequality is by Assumption \ref{ass:lipschitz_theta}. We then have
\begin{equation}
\lVert \nabla \bar{f}(\theta^{(m)}, {\lambda}) \rVert_2^2 \leq N^2 \bar{L}^2 (\lVert \lambda - \lambda^* \rVert_2 + \lVert \theta^{(m)} - \theta^* \rVert_2)^2 \leq M,
\end{equation}
because $ \theta^{(m)} $ are bounded by Assumption \ref{ass:bounded_para}.}
$ \qedwhite $

\begin{proposition}
	\label{pro:lipschitz_1}
	We have
	\begin{equation}
	f_i(X:\tilde{\theta}, \lambda) \leq f_i(X:\hat{\theta}, \lambda) + \nabla f_i(X:\hat{\theta}, \lambda)^T (\tilde{\theta}-\hat{\theta}) +\dfrac{1}{2}\bar{L} \lVert \tilde{\theta}-\hat{\theta} \rVert^2_2 , \forall \tilde{\theta},\hat{\theta},X.
	\label{res:lipschitz_continuity}
	\end{equation}
\end{proposition}
\textit{Proof.} This is a known result of the Lipschitz-continuous condition that can be found in \cite{Bottou2016}.
We have this result together with Assumption \ref{ass:lipschitz_theta}.


\subsection*{Proof of Theorem \ref{thm:lambda_converge}}

\begin{lemma}
	When 
	$  \sum_{m=1}^{\infty} \alpha^{(m)} <\infty  \text{  and  }
	\sum_{m=1}^{\infty} \sum_{n=1}^{m} \alpha^{(m)} \eta^{(n)} <\infty $,\\
	$ \tilde{\mu}_j^{(m)}:= \dfrac{\mu_j^{(m)}} {(1-\alpha^{(1)})(1-\alpha^{(2)})...(1-\alpha^{(m)})}  $ is a Cauchy series.
	\label{lemma:tilde_mu_converge}
\end{lemma}
\textit{Proof.} By Algorithm \ref{alg_BNIGA}, we have
\begin{equation}
\mu_j^{(m)} = \alpha^{(m)} \frac{1}{N} \sum_{i=1}^{N}{\cred a(W^{(m)}_{1,j,\cdot}X_{i})  }  + (1-\alpha^{(m)}) \mu_j^{(m-1)}.
\label{dfn:mum}
\end{equation}

We define 
$ \tilde{\alpha}^{(m)} := \dfrac{\alpha^{(m)}} {(1-\alpha^{(1)})(1-\alpha^{(2)})...(1-\alpha^{(m)})}$ 
and $\Delta W_{1,j,\cdot}^{(m)} := W_{1,j,\cdot}^{(m)} - W_{1,j,\cdot}^{(m-1)} $.  
After dividing \eqref{dfn:mum} by $ (1-\alpha^{(1)})(1-\alpha^{(2)})...(1-\alpha^{(m)}) $, we obtain
\begin{equation}
\tilde{\mu}_j^{(m)} = \tilde{\alpha}^{(m)}  \frac{1}{N} \sum_{i=1}^{N} {\cred a(W^{(m)}_{1,j,\cdot}X_{i})}  + \tilde{\mu}_j^{(m-1)}.
\label{dfn:mum2}
\end{equation}

Then we have 
\begin{equation*} 
\lvert \tilde{\mu}_j^{(m)} - \tilde{\mu}_j^{(m-1)} \rvert  
\leq \tilde{\alpha}^{(m)} \lvert {\cred k} \rvert \frac{1}{N} \sum_{i=1}^{N} 
\lvert \sum_{n=1}^{m} \Delta W^{(n)}_{1,j,\cdot}  X_{i} \rvert 
\label{stp:prf1_step2}
\end{equation*}

\begin{equation*} 
= \tilde{\alpha}^{(m)} \lvert {\cred k}  \rvert \frac{1}{N} \sum_{i=1}^{N} 
\left\lvert 
\sum_{n=1}^{m} \left(
\eta^{(n)}  \sum_{l=1}^{N} \nabla_{W_{1,j,\cdot}}f_l(
X_l:\theta^{(n)},\lambda^{(n)})
\right) \cdot X_{i} \right\rvert 
\end{equation*}

\begin{equation*} 
= \tilde{\alpha}^{(m)} \lvert {\cred k}  \rvert \frac{1}{N} \sum_{i=1}^{N} 
\sum_{n=1}^{m} 
\left(
\eta^{(n)}  
\left\lvert 
\left(
\sum_{l=1}^{N} \nabla_{W_{1,j,\cdot}}f_l(X_l:\theta^{(n)},\lambda^{(n)})
\right) \cdot X_{i} \right\rvert 
\right)
\end{equation*}


\begin{equation} 
\leq \tilde{\alpha}^{(m)} \lvert {\cred k}  \rvert \frac{1}{N} \sum_{i=1}^{N} \sum_{n=1}^{m}   
\left( 
\eta^{(n)}
\lVert   \sum_{l=1}^{N} \nabla_{W_{1,j,\cdot}}f_l(
X_l:\theta^{(n)},\lambda^{(n)}) 
\rVert
\cdot 
\lVert X_{i}  \rVert
\right)
\label{stp:prf1_step4}
\end{equation}

\begin{equation} \leq \tilde{\alpha}^{(m)} 
\lvert {\cred k}  \rvert  \sum_{i=1}^{N}
\sum_{n=1}^{m} \eta^{(n)}
\label{stp:prf1_step7}
\left(  \bar{L} \cdot  (\lVert W_{1,j,\cdot}^{(n)} - W_{1,j,\cdot}^* \rVert_2 +
\lVert \lambda_{j,\cdot}^{(n)} - \lambda_{j,\cdot}^* \rVert_2
)  \cdot \lVert X_{i} \rVert_2 \right) 
\end{equation}
\begin{equation} \leq \tilde{\alpha}^{(m)} 
\sum_{n=1}^{m} \left( \eta^{(n)} \right)
\lvert {\cred k}  \rvert  \sum_{i=1}^{N}
\left( 2 \bar{L} M \lVert X_{i} \rVert_2 \right) 
\label{stp:prf1_step8}
\end{equation}
\begin{equation} 
\leq \tilde{\alpha}^{(m)} \sum_{n=1}^{m}  \eta^{(n)} \tilde{M}_{\bar{L},M} .
\label{stp:prf1_step9}
\end{equation}

Equation \eqref{stp:prf1_step2} is due to 
$  W_{1,i,j}^{(m)} = \sum_{n=1}^{m} \Delta W_{1,i,j}^{(n)} . $

Therefore, 

\begin{equation}
\begin{aligned}
\lvert \tilde{\mu}_j^{(p)} - \tilde{\mu}_j^{(q)}  \rvert
\leq \tilde{M}_{\bar{L},M} \cdot  \sum_{m=p}^{q}\sum_{n=1}^{m} \tilde{\alpha}^{(m)}  \eta^{(n)} .
\end{aligned}
\label{mu_diff}
\end{equation}

%
%
%


It remains to show that
\begin{equation}
\sum_{m=1}^{\infty} \alpha^{(m)} <\infty,
\label{ass:1}
\end{equation}
\begin{equation}
\sum_{m=1}^{\infty} \sum_{n=1}^{m} \alpha^{(m)} \eta^{(n)} <\infty,
\label{ass:2}
\end{equation}

implies the convergence of $ \{\tilde{\mu}^{(m)}\} $.
By (\ref{ass:1}), we have
$ \Pi_{m=1}^{\infty} (1-\alpha^{(m)}) >0, $
since
$ \ln(\Pi_{m=1}^{\infty} (1-\alpha^{(m)}))  > \sum_{m=1}^{\infty} -\alpha^{(m)} > -\infty. $



It is also easy to show that there exists $ C $ and $ M_c $ such that for all $ m \geq M_c $, we have 
\begin{equation}
(1-\alpha^{(1)})(1-\alpha^{(2)})\dots(1-\alpha^{(m)}) \geq C .
\label{stp:prf_define_C}
\end{equation}

Therefore, $  \lim\limits_{m\rightarrow \infty} (1-\alpha^{(1)})(1-\alpha^{(2)})\dots(1-\alpha^{(m)}) \geq C . $

Thus the following holds:
\begin{equation} 
\tilde{\alpha}^{(m)}   \leq  \dfrac{1}{C}  {\alpha}^{(m)}
\label{ass:4_0} 
\end{equation}
and 
\begin{equation}
\sum_{m=p}^{q}\sum_{n=1}^{m} \tilde{\alpha}^{(m)}  \eta^{(n)} 
\leq 
\dfrac{1}{C} \sum_{m=p}^{q} \sum_{n=1}^{m}  {\alpha}^{(m)} \eta^{(n)} .
\label{ass:4} 
\end{equation}

From \eqref{ass:2} and \eqref{ass:4} it follows that the sequence $ \{\tilde{\mu}_j^{(m)}\} $ is a Cauchy series.  
$ \qedwhite $

\begin{lemma}
	Since $\{\tilde{\mu}_j^{(m)}\}   $ is a Cauchy series, $ \{\mu_j^{(m)} \} $ is a Cauchy series.
	\label{lemma:prf1_cauchy_for_mu}
\end{lemma}
\textit{Proof.}
We know that 
$  \mu_j^{(m)} = \tilde{\mu}_j^{(m)} (1-\alpha^{(1)}) ...(1-\alpha^{(m)}).  $
Since
$   \lim\limits_{m \rightarrow \infty} \tilde{\mu}_j^{(m)} \rightarrow \tilde{\mu}_j   $
and  $  \lim\limits_{m \rightarrow \infty} (1-\alpha^{(1)})...(1-\alpha^{(m)}) \rightarrow \tilde{C},  $
we have  $ \lim\limits_{m \rightarrow \infty} \mu_j^{(m)} \rightarrow \tilde{\mu}_j \cdot \tilde{C}.   $
Thus $ \mu_j^{(m)}  $ is a Cauchy series. 
$ \qedwhite $

\begin{lemma}
	If
	$  \sum_{m=1}^{\infty} \alpha^{(m)} <\infty  \text{  and  }
	\sum_{m=1}^{\infty} \sum_{n=1}^{m} \alpha^{(m)} \eta^{(n)} <\infty $,
	$ \{\sigma_j^{(m)}\} $ is a Cauchy series.
\end{lemma}
\textit{Proof.}
We define 
$  \sigma_j^{(m)} := \tilde{\sigma}_j^{(m)} (1-\alpha^{(1)}) ...(1-\alpha^{(m)})  $. Then we have
\begin{equation*}
\lvert \tilde{\sigma}_j^{(m+1)} - \tilde{\sigma}_j^{(m)} \rvert  
= \tilde{\alpha}^{(m)}
\sqrt{
	\frac{1}{N} \sum_{i=1}^{N} \left(  
	{\cred 
		a(W^{(m)}_{1,j,\cdot} X_{i}) }  - \mu_j^{(m)}
	\right)^2
}
\end{equation*}


\begin{equation}
= \tilde{\alpha}^{(m)} \dfrac{|{\cred k} |}{\sqrt{N}}
\sqrt{
	\sum_{i=1}^{N} \left( {\cred
		\dfrac{ a(W^{(m)}_{1,j,\cdot} X_{i}) }{k} } 
	- \dfrac{\mu_j^{(m)}}{{\cred k} }
	\right)^2
}.
\end{equation}

Since $ \{\mu_j^{(m)}\} $ is convergent, there exists $ c_1 $, $ c_2 $ and $ N_1 $ such that for any $ m > N_1 $, $ -\infty<c_1<\mu_j^{(m)}<c_2<\infty $.
For any $ \bar{C} \in 
\left\{\dfrac{c_1}{{\cred k} },  \dfrac{c_2}{{\cred k} }\right\}  $, we have 

\begin{equation}
\lvert \tilde{\sigma}_j^{(m+1)} - \tilde{\sigma}_j^{(m)} \rvert  
\leq \tilde{\alpha}^{(m)} \dfrac{{|{\cred k} |}}{\sqrt{N}}
\cdot
\sqrt{
	\sum_{i=1}^{N}  \left(  
	{\cred \dfrac{ a(W^{(m)}_{1,j,\cdot} X_{i}) }{k}} - \bar{C}
	\right)^2
}
\label{stp:prf2_step1}
\end{equation}

\begin{equation}
\leq \tilde{\alpha}^{(m)} \dfrac{{|{\cred k} |}}{\sqrt{N}}
\cdot
\sqrt{
	\sum_{i=1}^{N}  \left(  
	| {\cred \dfrac{ a(W^{(m)}_{1,j,\cdot} X_{i}) }{k}} |  + |\bar{C}|
	\right)^2
}
\label{stp:prf2_step1.5}
\end{equation}

\begin{equation}
\begin{aligned}
\leq \tilde{\alpha}^{(m)} \dfrac{{|{\cred k} |}}{\sqrt{N}}
\cdot 
\sqrt{
	\sum_{i=1}^{N}  \left(  
	\sum_{n=1}^{m}  \eta^{(n)} 
	\left(  2  N \bar{L}  M \lVert X_{i} \rVert_2 \right)
	+  \lvert\bar{C}\rvert
	\right)^2
}
\end{aligned}
\label{stp:prf2_step6}
\end{equation}

\begin{equation}
\leq \tilde{\alpha}^{(m)} \dfrac{{|{\cred k} |}}{\sqrt{N}}
\cdot 
\sqrt{ N \cdot
	\left(  
	\tilde{M}_{\bar{L},M} \sum_{n=1}^{m}\eta^{(n)} + \lvert \bar{C} \rvert
	\right)^2
}
\label{stp:prf2_step7}
\end{equation}

\begin{equation}
= \tilde{\alpha}^{(m)} |{\cred k} |
\cdot 
\left(  
\tilde{M}_{\bar{L},M} \sum_{n=1}^{m}\eta^{(n)} + \lvert \bar{C} \rvert
\right).
\label{stp:prf2_step9}
\end{equation}

Inequality \eqref{stp:prf2_step1.5} is by the following fact:
\begin{equation}
\sqrt{\sum_{i=1}^{n} (a_i-c)^2 } 
\leq 
\sqrt{\sum_{i=1}^{n} (|a_i|+|c|)^2 } ,
\label{stp:prf2_add_1}
\end{equation}

where $ b $ and $ a_i$ for every $ i $ are arbitrary real scalars. Besides, \eqref{stp:prf2_add_1} is due to 
$ -2  a_i c \leq \max 
\{-2 |a_i| c, 
2  |a_i|  c\}. $

Inequality \eqref{stp:prf2_step6} follow from the square function being increasing for nonnegative numbers.
Besides these facts, \eqref{stp:prf2_step6} is also by the same techniques we used in \eqref{stp:prf1_step4}-\eqref{stp:prf1_step8} where we bound the derivatives with the Lipschitz continuity in the following inequality:
\begin{equation}
\lVert
\sum_{l=1}^{N} \nabla_{W_{1,j,\cdot}}f_l(
X_l:\theta^{(n)},\lambda^{(n)})
\rVert
\leq
2  N \bar{L} M  .
\end{equation} 

Inequality \eqref{stp:prf2_step7} is by collecting the bounded terms into a single bound $ \tilde{M}_{\bar{L},M}$.
Therefore, 
\begin{equation}
\label{stp:prf1_converge_sigma}
\lvert \tilde{\sigma}_j^{(q)} - \tilde{\sigma}_j^{(p)} \rvert   
\leq  
\sum_{m=p}^{q-1}
\tilde{\alpha}^{(m)} | {\cred k} |
\cdot 
\left(  
\tilde{M}_{\bar{L},M} \sum_{n=1}^{m}\eta^{(n)} + |\bar{C}|
\right).
\end{equation}

Using the similar methods in deriving \eqref{ass:1} and \eqref{ass:2}, it can be seen that a set of sufficient conditions ensuring the convergence for $
\{\tilde{\sigma}_j^{(m)}\} $ is:
$ \sum_{m=1}^{\infty} \alpha^{(m)} <\infty $
and 
$ \sum_{m=1}^{\infty}  \sum_{n=1}^{m} \alpha^{(m)}  \eta^{(n)}   <\infty. $

Therefore, the convergence conditions for $ \{{\sigma}_j^{(m)}\} $ are the same as for $ \{\mu_j^{(m)}\} $. $ \qedwhite $

It is clear that these lemmas establish the proof of Theorem \ref{thm:lambda_converge}.

\subsection*{Consequences of Theorem \ref{thm:lambda_converge}}

\begin{proposition}
	\label{pro:lambda_upper_bound}
	Under the assumptions of Theorem \ref{thm:lambda_converge}, we have
$ 	\lvert \lambda^{(m)} - \bar{\lambda} \rvert_{\infty} \leq a_m , $
	where
	\begin{equation}
	a_m = M_1 \sum_{i=m}^{\infty}  \sum_{j=1}^{i} {\alpha}^{(i)} \eta^{(j)}
	+ 
	M_2 \sum_{i=m}^{\infty} \alpha^{(i)} .
	\label{dfn:a_m}
	\end{equation}
	$ M_1 $ and $ M_2 $ are constants.
\end{proposition}
\textit{Proof.} 
For the upper bound of $ \sigma_j^{(m)}$, by \eqref{stp:prf2_step9}, we have
\begin{equation*}
\lvert \tilde{\sigma}_j^{(q)} - \tilde{\sigma}_j^{(p)} \rvert   
\leq  
\sum_{m=p}^{q-1}
\tilde{\alpha}^{(m)} | {\cred k} |
\left(  
\tilde{M}_{\bar{L},M} \sum_{n=1}^{m}\eta^{(n)} + |\bar{C}|
\right).
\end{equation*}
We define $ \tilde{\sigma}_j   := \dfrac{\bar{\sigma}_j }{(1-\alpha^{(1)})...(1-\alpha^{(u)})...} $.
Therefore,
\begin{equation}
\begin{aligned}
\lvert \tilde{\sigma}_j  - \tilde{\sigma}_j^{(m)} \rvert   
&\leq  
\sum_{i=m}^{\infty}
\tilde{\alpha}^{(i)} |{\cred k} |
\left(  
\tilde{M}_{\bar{L},M} \sum_{j=1}^{i}\eta^{(j)} + |\bar{C}|
\right)  \\
&\leq
\dfrac{|{\cred k} |}{{C}}
\sum_{i=m}^{\infty}
{\alpha}^{(i)} 
\left(  
\tilde{M}_{\bar{L},M} \sum_{j=1}^{i}\eta^{(j)} + |\bar{C}|
\right).
\end{aligned}
\label{sigma_tilde_diff_step_bar}
\end{equation}

The first inequality comes by substituting $ p $ by $ m $ and by taking $ \lim $ as  $q \rightarrow \infty$ in \eqref{stp:prf1_converge_sigma}. 
The second inequality comes from \eqref{stp:prf_define_C}.
We then obtain, 
\begin{equation}
\begin{aligned}
&\left\lvert
{\sigma_j^{(m)} - \bar{\sigma}_j}\
\right\rvert 
\leq
\left\lvert
\tilde{\sigma}_j^{(m)}
-
\tilde{\sigma}_j^{(\infty)}
\right\rvert
+
\left\lvert
\dfrac{\bar{\sigma}_j }{(1-\alpha^{(1)})...(1-\alpha^{(m)})}
-
\tilde{\sigma}_j^{(\infty)}
\right\rvert \\
=&  \left\lvert
\tilde{\sigma}_j^{(m)}
-
\tilde{\sigma}_j^{(\infty)}
\right\rvert
+
\left\lvert 
\dfrac{ \bar{\sigma}_j }{(1-\alpha^{(1)})...(1-\alpha^{(m)})}
- \dfrac{ \bar{\sigma}_j }{(1-\alpha^{(1)})...(1-\alpha^{(u)})...}\right\rvert \\
=&
\left\lvert
\tilde{\sigma}_j^{(m)}
-
\tilde{\sigma}_j^{(\infty)}
\right\rvert
+
\bar{\sigma}_j 
\left\lvert 
\dfrac{ (1-\alpha^{(m+1)})...(1-\alpha^{(u)})... - 1 }{(1-\alpha^{(1)})...(1-\alpha^{(u)})...}\right\rvert \\
\leq&
\left\lvert
\tilde{\sigma}_j^{(m)}
-
\tilde{\sigma}_j^{(\infty)}
\right\rvert
+
\dfrac{\bar{\sigma}_j }{{C}} 
\lvert 
{ 1-(1-\alpha^{(m+1)})...(1-\alpha^{(u)})...}\rvert \\
\leq & \left\lvert
\tilde{\sigma}_j^{(m)}
-
\tilde{\sigma}_j^{(\infty)}
\right\rvert+ 
\dfrac{\bar{\sigma}_j }{{C}} \sum_{n=m+1}^{\infty} \alpha^{(n)}.
\label{sigma_diff_step_true}
\end{aligned}
\end{equation}
The second inequality is by $ (1-\alpha^{(1)})...(1-\alpha^{(m)}) <1 $,
the third inequality is by \eqref{stp:prf_define_C} and
the last inequality can be easily seen by induction. By \eqref{sigma_diff_step_true}, we obtain
\begin{equation}
\lvert \bar{\sigma}_j - {\sigma}_j^{(m)} \rvert   
=\lim\limits_{M\rightarrow \infty} \lvert {\sigma}_j^{(M)} - {\sigma}_j^{(m)} \rvert    
\leq
\lvert \tilde{\sigma}_j  - \tilde{\sigma}_j^{(m)} \rvert   + \dfrac{ \bar{\sigma_j} }{C} \sum_{n=m+1}^{\infty} \alpha^{(n)}.
\label{sigma_diff_step_true_2}
\end{equation}
Therefore, we have
\begin{equation}
\begin{aligned}
&\lvert \bar{\sigma}_j - {\sigma}_j^{(m)} \rvert   \leq 
\lvert \tilde{\sigma}_j  - \tilde{\sigma}_j^{(m)} \rvert   + \dfrac{\bar{\sigma}_j}{C} \sum_{n=m+1}^{\infty} \alpha^{(n)} \\
\leq & 
\sum_{i=m}^{\infty}
\tilde{\alpha}^{(i)} | {\cred k} |
\cdot 
\left(  
\tilde{M}_{\bar{L},M} \sum_{j=1}^{i}\eta^{(j)} + |\bar{C}|
\right)
+ \dfrac{\bar{\sigma}_j}{C} \sum_{i=m+1}^{\infty} \alpha^{(i)} \\
\leq &
\sum_{i=m}^{\infty}
\dfrac{1}{C}
{\alpha}^{(i)} | {\cred k} |
\cdot 
\left(  
\tilde{M}_{\bar{L},M} \sum_{j=1}^{i}\eta^{(j)} + |\bar{C}|
\right)
+ \dfrac{\bar{\sigma}_j}{C} \sum_{i=m+1}^{\infty} \alpha^{(i)}  \\
\leq &
\dfrac{\tilde{M}_{\bar{L},M} | {\cred k} |}{C}
\sum_{i=m}^{\infty}
\sum_{j=1}^{i}
{\alpha}^{(i)} \eta^{(j)}
+ 
\left(
\dfrac{\bar{\sigma}_j}{C} 
+
\dfrac{| {\cred k} ||\bar{C}|}{C}
\right)
\sum_{i=m}^{\infty} \alpha^{(i)} .
\label{prf:state_sigma_convergence}
\end{aligned}  
\end{equation}
The first inequality is by \eqref{sigma_diff_step_true_2}, the second inequality is by \eqref{stp:prf1_converge_sigma}, the third inequality is by \eqref{ass:4_0} and the fourth inequality is by adding the nonnegative term $ \dfrac{\bar{\sigma}_j}{C} \alpha^{(m)} $  to the right-hand side.

For the upper bound of $ \mu_j^{(m)},$ we have
\begin{equation}
\begin{aligned}
&\left\lvert
{\mu_j^{(m)} -\bar{\mu}_j}\
\right\rvert 
\leq 
\left\lvert
\tilde{\mu}^{(m)}
-
\tilde{\mu}^{(\infty)}
\right\rvert
+
\left\lvert
\dfrac{\bar{\mu}_j}{(1-\alpha^{(1)})...(1-\alpha^{(m)})}
-
\tilde{\mu}^{(\infty)}
\right\rvert.
\end{aligned}
\label{stp:res_no_tilde_step_to_tilde}
\end{equation}
Let us define $ A_m := \left\lvert \tilde{\mu}^{(m)} - \tilde{\mu}^{(\infty)}
\right\rvert $ and 
$ B_m := \left\lvert
\dfrac{\bar{\mu}_j}{(1-\alpha^{(1)})...(1-\alpha^{(m)})}
-
\tilde{\mu}^{(\infty)}
\right\rvert $. 
Recall from Theorem \ref{thm:lambda_converge} that $ \{\mu_j^{(m)}\} $ is a Cauchy series, by \eqref{mu_diff}, 
$ \lvert \tilde{\mu}_j^{(p)} - \tilde{\mu}_j^{(q)}  \rvert
\leq   
\bar{M}_{\bar{L},M} \cdot  \sum_{m=p}^{q}\sum_{n=1}^{m} {\alpha}^{(m)}  \eta^{(n)} . $
Therefore, the first term in \eqref{stp:res_no_tilde_step_to_tilde} is bounded by
\begin{equation}
\lvert \tilde{\mu}_j^{(m)} - \tilde{\mu}_j^{\infty}  \rvert
\leq   
\tilde{M}_{\bar{L},M} \cdot  \sum_{i=m}^{\infty}\sum_{n=1}^{i} {\alpha}^{(i)}  \eta^{(n)}  < \infty.
\label{prf:stp_mu_tilde_convergence_bound}
\end{equation}

For the second term in \eqref{stp:res_no_tilde_step_to_tilde}, recall that $ C := (1-\alpha^{(1)})...(1-\alpha^{(u)})...$.
Then we have
$ \begin{aligned}
&C \cdot \left\lvert
\dfrac{\bar{\mu}_j}{(1-\alpha^{(1)})...(1-\alpha^{(m)})}
-
\tilde{\mu}^{(\infty)}
\right\rvert \leq \bar{\mu}_j \sum_{i=m+1}^{\infty} \alpha^{(i)},
\end{aligned} $
where the inequality can be easily seen by induction.
Therefore, the second term in \eqref{stp:res_no_tilde_step_to_tilde} is bounded by
\begin{equation}
\left\lvert
\dfrac{\bar{\mu}_j}{(1-\alpha^{(1)})...(1-\alpha^{(m)})}
-
\tilde{\mu}^{(\infty)}
\right\rvert
\leq
\dfrac{ \bar{\mu}_j}{C}
\sum_{i=m+1}^{\infty} \alpha^{(i)}.
\label{prf:stp_B_m_term_bound_in_mu_convergence}
\end{equation}
From these we obtain
\begin{equation}
\begin{aligned}
& \left\lvert
{\mu_j^{(m)} -\bar{\mu}_j}\
\right\rvert 
\leq 
\tilde{M}_{\bar{L},M} \sum_{i=m}^{\infty}\sum_{n=1}^{i} {\alpha}^{(i)}  \eta^{(n)}
+
\dfrac{ \bar{\mu}_j}{C}
\sum_{i=m+1}^{\infty} \alpha^{(i)}.
\label{prf:state_mu_convergence}
\end{aligned}
\end{equation}
The first inequality is by \eqref{stp:res_no_tilde_step_to_tilde} and the second inequality is by \eqref{prf:stp_mu_tilde_convergence_bound} and \eqref{prf:stp_B_m_term_bound_in_mu_convergence}.  Combining \eqref{prf:state_sigma_convergence} and \eqref{prf:state_mu_convergence}, we have that 
\begin{equation*}
\lvert \lambda^{(m)} - \bar{\lambda} \rvert_{\infty} 
\leq
M_1
\sum_{i=m}^{\infty}
\sum_{j=1}^{i}
{\alpha}^{(i)} \eta^{(j)}
+ 
M_2
\sum_{i=m}^{\infty} \alpha^{(i)},
\end{equation*}
where $ M_1 $ and $ M_2 $ are constants defined as 
$   M_1 = \max(\dfrac{\tilde{M}_{\bar{L},M} | {\cred k} |}{C}, \bar{M}_{\bar{L},M})   $
and 
$   M_2 = \max(\dfrac{\bar{\sigma}_j+ | {\cred k} ||\bar{C}|}{C}, \dfrac{ \bar{\mu}_j}{C}).  \qedwhite $

\begin{proposition}
	\label{pro:gradient_bounded}
	Under the assumptions of Theorem \ref{thm:lambda_converge},
	\[  - \nabla \bar{f} (\theta^{(m)}, \bar{\lambda})^T \cdot \nabla \bar{f} (\theta^{(m)}, \lambda^{(m)}) \leq - \lVert \nabla \bar{f} (\theta^{(m)}, \bar{\lambda}) \rVert^2 + \bar{L} M \sqrt{n_2} a_m,   \]
	where $ a_m $ is defined in Proposition \ref{pro:lambda_upper_bound}. 
\end{proposition}
\textit{Proof.} For simplicity of the proof, let us define 
\label{prf:section:convergence_of_gradeint}
$   x^{(m)} := \nabla \bar{f} (\theta^{(m)}, \bar{\lambda}), \quad  y^{(m)} := \nabla \bar{f} (\theta^{(m)}, \lambda^{(m)})  . $
We have
\begin{equation}
\begin{aligned}
\lvert x^{(m)} - y^{(m)} \rvert_\infty  
\leq  \bar{L} \sqrt{n_2} \lVert \lambda^{(m)} - \bar{\lambda} \rVert_\infty 
\leq  \bar{L} \sqrt{n_2} a_m,
\end{aligned}
\label{prf:pro:84:1}
\end{equation}
where $ \sqrt{n_2} $ is the dimension of $ \lambda $.
The second inequality is by Assumption \ref{ass:lipschitz_theta} and the fourth inequality is by Proposition \ref{pro:lambda_upper_bound}. 
Inequality \eqref{prf:pro:84:1} implies that for all $ m $ and $ i $, we have
$ \lvert x^{(m)}_i - y^{(m)}_i \rvert \leq \bar{L} \sqrt{n_2} a_m . $

It remains to show 
\begin{equation}
- \sum_{i}y^{(m)}_i x^{(m)}_i 
\leq - \sum_{i} {x^{(m)}_i}^2 + \bar{L}M \sqrt{n_2} a_m 
,\forall i, m.
\label{prf:extra1_statement1}
\end{equation}
This is established by the following four cases.

1) If $ {x^{(m)}_i} \geq 0, {x^{(m)}_i} - {y^{(m)}_i} \geq 0  $, then $ {x^{(m)}_i} \leq \bar{L}\sqrt{n_2}a_m + {y^{(m)}_i}  $. Thus $ -{x^{(m)}_i} {y^{(m)}_i} \leq -{x^{(m)}_i}^2 + \bar{L}M \sqrt{n_2}a_m $ by Proposition \ref{pro:upper_bound_norm_gradient}.

2) If $ {x^{(m)}_i} \geq 0, {x^{(m)}_i} - {y^{(m)}_i} \leq 0  $, then $ {x^{(m)}_i} \leq  {y^{(m)}_i}  $, $ {x^{(m)}_i}^2 \leq {x^{(m)}_i} \cdot {y^{(m)}_i}$ and $ -{x^{(m)}_i} {y^{(m)}_i} \leq -{x^{(m)}_i}^2 .$

3) If $ {x^{(m)}_i} < 0, {x^{(m)}_i} - {y^{(m)}_i} \geq 0  $, then $ {x^{(m)}_i} \geq  {y^{(m)}_i}  $, $ {x^{(m)}_i}^2 \leq {x^{(m)}_i} \cdot {y^{(m)}_i}$ and $ -{x^{(m)}_i} {y^{(m)}_i} \leq -{x^{(m)}_i}^2 .$

4) If $ {x^{(m)}_i} < 0, {x^{(m)}_i} - {y^{(m)}_i} \leq 0  $, then $ {y^{(m)}_i} - {x^{(m)}_i} \leq \bar{L} \sqrt{n_2} a_m  $, $ {y^{(m)}_i} {x^{(m)}_i} - {x^{(m)}_i}^2 \geq \bar{L}\sqrt{n_2} a_m {x^{(m)}_i}  $ and $ -{y^{(m)}_i} {x^{(m)}_i} \leq -{x^{(m)}_i}^2  - \bar{L} \sqrt{n_2} a_m {x^{(m)}_i} \leq -{x^{(m)}_i}^2 + \bar{L}M \sqrt{n_2}a_m $. The last inequality is by Proposition \ref{pro:upper_bound_norm_gradient}.  

All these four cases yield \eqref{prf:extra1_statement1}. $ \qedwhite $


\begin{proposition}
	\label{pro:lipschitz_2} 
	Under the assumptions of Theorem \ref{thm:lambda_converge}, we have 
	
	\begin{equation}
	\begin{aligned}
	 \bar{f}(\theta^{(m+1)}, \bar{\lambda} )  \leq &    \bar{f}(\theta^{(m)} , \bar{\lambda} )  - \eta^{(m)} \lVert \nabla \bar{f}(\theta^{(m)},\bar{\lambda})   \rVert^2_2
	\\
	 & 	+ \eta^{(m)} \bar{L} M \sqrt{n_2} a_m
	+  \dfrac{1}{2} (\eta^{(m)})^2 \cdot   N\bar{L}M,
	\end{aligned}
	\label{res:lipschitz_continuity_2}
	\end{equation}
	where $ M $ is a constant and $ a_m $ is defined in Proposition \ref{pro:lambda_upper_bound}. 
\end{proposition}
\textit{Proof.} By Proposition \ref{pro:lipschitz_1}, 
\begin{equation*}
f_i(X_i:\tilde{\theta}, {\lambda}) \leq 
f_i(X_i:\hat{\theta}, {\lambda}) + \nabla f_i(X_i:\hat{\theta}, {\lambda})^T(\tilde{\theta}-\hat{\theta}) + \dfrac{1}{2} \bar{L} \lVert \tilde{\theta}-\hat{\theta} \rVert_2^2.
\end{equation*}
Therefore, we can sum it over the entire training set from $ i=1 $ to $ N $ to obtain
\begin{equation}
\bar{f}(\tilde{\theta}, {\lambda}) \leq 
\bar{f}(\hat{\theta}, {\lambda}) + \nabla \bar{f}(\hat{\theta}, {\lambda})^T(\tilde{\theta}-\hat{\theta}) 
+ \dfrac{N}{2} \bar{L} \lVert \tilde{\theta}-\hat{\theta} \rVert_2^2.
\label{stp:prf_10_f_bar}
\end{equation}

In Algorithm \ref{alg_BNIGA}, we define the update of $ \theta $ in the following full gradient way:
\begin{equation}
\theta^{(m+1)}  := \theta^{(m)} - \eta^{(m)} \cdot \sum_{i=1}^{N} \cdot \nabla  f_i(X_i: \theta^{(m)}, {\lambda^{(m)}}), 
\end{equation}
which implies
\begin{equation}
\theta^{(m+1)} - \theta^{(m)} =  - \eta^{(m)} \cdot  \nabla \bar{f}(\theta^{(m)}, {\lambda^{(m)}}).
\label{stp:prf_10_full_gradient_update}
\end{equation}

By \eqref{stp:prf_10_full_gradient_update} we have 
$  \tilde{\theta} - \hat{\theta} =  \theta^{(m+1)}  -  \theta^{(m)}  = -\eta^{(m)} \nabla \bar{f}(\theta^{(m)}, {\lambda^{(m)}})  .$ We now substitute $ \tilde{\theta} := \theta^{(m+1)} $, $ \hat{\theta}:=\theta^{(m)} $ and $ \lambda:=\bar{\lambda} $ into \eqref{stp:prf_10_f_bar} to obtain
\begin{equation}
\begin{aligned}
& \bar{f}(\theta^{(m+1)}, {\bar{\lambda}} )  \\ 
 \leq  &
\bar{f}(\theta^{(m)}, {\bar{\lambda}} ) - \eta^{(m)} \nabla \bar{f}(\theta^{(m)}, {\bar{\lambda}})^T \nabla \bar{f}(\theta^{(m)}, {\lambda^{(m)}})  +  (\eta^{(m)})^2 \cdot \dfrac{N\bar{L}M}{2} \\
 \leq &
\bar{f}(\theta^{(m)}, {\bar{\lambda}} ) - \eta^{(m)} \lVert \nabla \bar{f}(\theta^{(m)}, {\bar{\lambda}}) \rVert^2_2 
+ \eta^{(m)} \bar{L} M \sqrt{n_2} a_m  \\
 & + \dfrac{1}{2}  (\eta^{(m)})^2 \cdot N\bar{L}M  .
\end{aligned}
\label{prf:bound_diff_m+1_m_f}
\end{equation}

The first inequality is by plugging \eqref{stp:prf_10_full_gradient_update} into \eqref{stp:prf_10_f_bar}, {\cb the second inequality comes from Proposition \ref{pro:upper_bound_norm_gradient} }and the third inequality comes from Proposition \ref{pro:gradient_bounded}.  $ \qedwhite $

\subsection*{Proof of Theorem \ref{thm:double_limits}}
Here we show Theorem \ref{thm:double_limits} as the consequence of Theorem \ref{thm:lambda_converge} and Lemmas \ref{thm:nonconvex_diminishing2}, \ref{thm:nonconvex_diminishing} and \ref{lem:nonconvex_limit_gradient_zero}.

\subsubsection*{Proof of Lemma \ref{thm:nonconvex_diminishing2}}

Here we show Lemma \ref{thm:nonconvex_diminishing2} as the consequence of Lemmas \ref{lem:sum_sigma_infty_m_finite}, \ref{lemma:gap_bar_f_lambda_m_bar_finite} and \ref{lem:nonconvex_dimnishing2_last_lemma}.
\begin{lemma}
	\label{lem:sum_sigma_infty_m_finite}
$ 	\sum_{m=1}^{\infty} \sum_{i=m}^{\infty} \sum_{n=1}^{i} \alpha^{(i)} \eta^{(n)} <\infty $
	and 
$ 	\sum_{m=1}^{\infty} \sum_{n=m}^{\infty} \alpha^{(n)} <\infty $
	is a set of sufficient condition to ensure
	\begin{equation}
	\sum_{m=1}^{\infty} \lvert \bar{\sigma_j} -  \sigma_j^{(m)} \rvert <\infty , \forall j.
	\label{lem:eqn:sum_sigma_infty_m_finite}
	\end{equation}
\end{lemma}
\textit{Proof.}
By plugging \eqref{sigma_diff_step_true_2} and \eqref{sigma_tilde_diff_step_bar} into \eqref{lem:eqn:sum_sigma_infty_m_finite}, we have the following for all $ j $:
\begin{equation}
\begin{aligned}
& \sum_{m=1}^{\infty}
{ \left\lvert \bar{\sigma}_j  - \sigma^{(m)}_j \right\rvert} 
\leq 
\sum_{m=1}^{\infty}
\left(
\lvert \tilde{\sigma}_j  - \tilde{\sigma}_j^{(m)} \rvert  
+   \dfrac{\bar{\sigma}_j }{{C}}   \sum_{n=m+1}^{\infty} \alpha^{(n)}
\right) \\
\leq & 
\dfrac{|{\cred k} |\cdot   \tilde{M}_{\bar{L},M}}{C}
\sum_{m=1}^{\infty}
\sum_{i=m}^{\infty}
{\alpha}^{(i)} 
\sum_{j=1}^{i}\eta^{(j)}
+ 
\dfrac{  {\bar{\sigma}_j } + |{\cred k} ||\bar{C}|}{C}
\sum_{m=1}^{\infty}
\sum_{n=m+1}^{\infty} \alpha^{(n)}.
\end{aligned}
\label{stp:prf3_4}
\end{equation}

It is easy to see that the the following conditions are sufficient for right-hand side of \eqref{stp:prf3_4} to be finite:
$ \sum_{m=1}^{\infty} \sum_{i=m}^{\infty} \sum_{n=1}^{i} \alpha^{(i)} \eta^{(n)} <\infty $
and 
$ \sum_{m=1}^{\infty} \sum_{n=m}^{\infty} \alpha^{(n)} <\infty. $

Therefore, we obtain
\hfill
$ \displaystyle \sum_{m=1}^{\infty} \lvert \bar{\sigma_j} -  \sigma_j^{(m)} \rvert <\infty , \forall j.  $ $ \qedwhite $

\begin{lemma} 
	\label{lemma:gap_bar_f_lambda_m_bar_finite}
	Under Assumption \ref{ass:l_lipschitz}, 
	\[ \sum_{m=1}^{\infty} \sum_{i=m}^{\infty} \sum_{n=1}^{i} \alpha^{(i)} \eta^{(n)} <\infty \quad \text{and} \quad  
	\sum_{m=1}^{\infty} \sum_{n=m}^{\infty} \alpha^{(n)} <\infty \]
	is a set of sufficient conditions to ensure 
	\[ \limsup\limits_{M\rightarrow \infty}  \sum_{m=1}^{M}
	\left|
	\bar{f}(\theta^{(m)},\lambda^{(m)}) - \bar{f}(\theta^{(m)}, \bar{\lambda})  
	\right|
	<\infty.  \]
\end{lemma}
\textit{Proof.}
By Assumption \ref{ass:l_lipschitz}, we have 
\begin{equation} \label{res:ass_l_lipschitz}
\lVert l_i(x) - l_i(y)\rVert 
\leq \hat{M} \lVert x - y \rVert 
\leq \hat{M} \sum_{i=1}^{D} \lvert x_i - y_i \rvert.
\end{equation}
By the definition of $ f_i(\cdot) $, we then have 
\begin{align}
\allowdisplaybreaks[4]
& \sum_{m=1}^{\infty}
\left|
\bar{f}(\theta^{(m)},\lambda^{(m)}) - \bar{f}(\theta^{(m)}, \bar{\lambda})  
\right| \\
\leq &
\sum_{m=1}^{\infty} \sum_{i=1}^{N}
\left\lvert
\left( l_i(X_i:\theta^{(m)},\lambda^{(m)}) - l_i(X_i:\theta^{(m)},\bar{\lambda})\right)
\right\rvert \\
\leq &
{ \cred M_2 \sum_{m=1}^{\infty} \sum_{j=1}^{D}  \sum_{i=1}^{N}
	\left\lvert
	\frac{   a(W^{(m)}_{1, j,\cdot} X_i) -\mu^{(m)}_j}{\sigma^{(m)}_j+ \epsilon_B}  
	-     \frac{  a(W^{(m)}_{1, j,\cdot} X_i) -\bar{\mu}_j}{\bar{\sigma}_j+ \epsilon_B}
	\right\rvert}  \\
\leq &
M_3
\sum_{m=1}^{\infty}
\sum_{j=1}^{D}
\left(
\sum_{i=1}^{N}
|{\cred k}  |
|W^{(m)}_{1, j,\cdot} X_i| 
\left\lvert
\frac{ \bar{\sigma}_j - \sigma^{(m)}_j}{\epsilon_B^2}
\right\rvert
+
N
\left\lvert
\frac{  \bar{\mu}_j}{\bar{\sigma}_j+ \epsilon_B}      
- \frac{  \mu^{(m)}_j}{\sigma^{(m)}_j+ \epsilon_B}
\right\rvert
\right).
\label{prf:2_4_6_true}
\end{align}


The first inequality is by the Cauchy-Schwarz inequality, and the second one is by \eqref{res:ass_l_lipschitz}.
To show the finiteness of \eqref{prf:2_4_6_true}, we only need to show the following two statements:

\begin{equation}
\sum_{m=1}^{\infty}
\sum_{i=1}^{N}
|{\cred k}  |
|W^{(m)}_{1, j,\cdot}  X_i| 
\left\lvert
\frac{ \bar{\sigma}_j - \sigma^{(m)}_j}{\epsilon_B^2}
\right\rvert
< \infty , \forall j
\label{stp:prf2_finite_part2}
\end{equation}
and
\begin{equation}
\sum_{m=1}^{\infty}
\left\lvert
\frac{  \bar{\mu}_j}{\bar{\sigma}_j+ \epsilon_B}      
- \frac{  \mu^{(m)}_j}{\sigma^{(m)}_j+ \epsilon_B}
\right\rvert
< \infty , \forall j.
\label{stp:prf2_finite_part1}
\end{equation}

\textit{Proof of \eqref{stp:prf2_finite_part2}:} 
For all $ j $ we have 

\begin{equation}
\begin{aligned}
&\sum_{m=1}^{\infty}
\sum_{i=1}^{N}
|{\cred k}  |
|W^{(m)}_{1, j,\cdot}  X_i| 
\left\lvert
\frac{ \bar{\sigma}_j - \sigma^{(m)}_j}{\epsilon_B^2}
\right\rvert \\
\leq &
\sum_{m=1}^{\infty}
|{\cred k} | NDM \max\limits_{i}\lVert X_i \rVert
\dfrac{1}{\epsilon_B^2}
\left\lvert
\bar{\sigma}_j - \sigma^{(m)}_j
\right\rvert \\
= &
|{\cred k} | NDM \max\limits_{i}\lVert X_i \rVert
\dfrac{1}{\epsilon_B^2}
\sum_{m=1}^{\infty}
\left\lvert
\bar{\sigma}_j - \sigma^{(m)}_j
\right\rvert.
\end{aligned}
\label{stp:prf2_finite_part3}
\end{equation}

{\cb The inequality comes from $ |W^{(m)}_{1, j,\cdot}  X_i| \leq D M \lVert X_i \rVert_2 $, where $ D $ is the dimension of $ X_i $ and $ M $ is the element-wise upper bound for $ W^{(m)}_{1, j,\cdot} $ in Assumption \ref{ass:bounded_para}.} 

Finally, we invoke Lemma \ref{lemma:tilde_mu_converge} to assert that
$ \sum_{m=1}^{\infty}
\left\lvert
\bar{\sigma}_j - \sigma^{(m)}_j
\right\rvert $ is finite.

\textit{Proof of \eqref{stp:prf2_finite_part1}:} For all $ j $ we have 
\begin{equation}
\label{prf:2_4_3}
\begin{aligned}
&\sum_{m=1}^{\infty}
\left\lvert
\frac{  \bar{\mu}_j}{\bar{\sigma}_j+ \epsilon_B}      
- \frac{  \mu^{(m)}_j}{\sigma^{(m)}_j+ \epsilon_B}
\right\rvert \\
\leq &
\sum_{m=1}^{\infty}
\left\lvert
\frac{  \bar{\mu}_j}{\bar{\sigma}_j+ \epsilon_B}      
- \frac{  \mu^{(m)}_j}{\bar{\sigma}_j+ \epsilon_B}      
\right\rvert 
+
\sum_{m=1}^{\infty}
\left\lvert
\frac{  \mu^{(m)}_j}{\bar{\sigma}_j+ \epsilon_B}      
- \frac{  \mu^{(m)}_j}{\sigma^{(m)}_j+ \epsilon_B}
\right\rvert .
\end{aligned}
\end{equation}

The first term in \eqref{prf:2_4_3} is finite since $ \{\mu_j^{(m)}\} $ is a Cauchy series. For the second term, we know that there exists a constant $ M $ such that for all $ m \geq M $, $ \mu^{(m)}_j \leq \bar{\mu} +1. $ This is also by the fact that $ \{\mu_j^{(m)}\} $ is a Cauchy series and it converges to $ \bar{\mu} $. Therefore, the second term in  \eqref{prf:2_4_3} becomes
\begin{equation}
\begin{aligned}
&\sum_{m=1}^{M-1}
\left\lvert
\frac{  \mu^{(m)}_j}{\bar{\sigma}_j+ \epsilon_B}      
- \frac{  \mu^{(m)}_j}{\sigma^{(m)}_j+ \epsilon_B}
\right\rvert 
+
\sum_{m=M}^{\infty}
\left\lvert
\frac{  \mu^{(m)}_j}{\bar{\sigma}_j+ \epsilon_B}      
- \frac{  \mu^{(m)}_j}{\sigma^{(m)}_j+ \epsilon_B}
\right\rvert 
\\
\leq & 
\sum_{m=1}^{M-1}
\left\lvert
\frac{  \mu^{(m)}_j}{\bar{\sigma}_j+ \epsilon_B}      
- \frac{  \mu^{(m)}_j}{\sigma^{(m)}_j+ \epsilon_B}
\right\rvert 
+
\sum_{m=M}^{\infty}
(\bar{\mu}+1)
\left\lvert
\frac{  1}{\bar{\sigma}_j+ \epsilon_B}      
- \frac{  1}{\sigma^{(m)}_j+ \epsilon_B}
\right\rvert .
\label{prf:2_4_4}
\end{aligned}
\end{equation}

Noted that function $ f(\sigma) = \dfrac{1}{\sigma+\epsilon_B} $ is Lipschitz continuous since its gradient is bounded by $ \dfrac{1}{\epsilon_B^2}$. Therefore we can choose $ \dfrac{1}{\epsilon_B^2}$ as the Lipschitz constant for $ f(\sigma) $. We then have the following inequality:

\begin{equation}
\left\lvert
\frac{  1}{\bar{\sigma}_j+ \epsilon_B}      
- \frac{  1}{\sigma^{(m)}_j+ \epsilon_B}
\right\rvert
\leq
\dfrac{1}{\epsilon_B^2} \lvert \bar{\sigma}_j - \sigma^{(m)}_j\rvert.
\label{stp:prf3_lipschitz_f}
\end{equation}

Plugging \eqref{stp:prf3_lipschitz_f} into \eqref{prf:2_4_4}, we obtain
\begin{equation}
\begin{aligned}
& 
\sum_{m=1}^{M-1}
\left\lvert
\frac{  \mu^{(m)}_j}{\bar{\sigma}_j+ \epsilon_B}      
- \frac{  \mu^{(m)}_j}{\sigma^{(m)}_j+ \epsilon_B}
\right\rvert 
+
\sum_{m=M}^{\infty}
(\bar{\mu}+1)
\left\lvert
\frac{  1}{\bar{\sigma}_j+ \epsilon_B}      
- \frac{  1}{\sigma^{(m)}_j+ \epsilon_B}
\right\rvert \\
\leq &
\sum_{m=1}^{M-1}
\left\lvert
\frac{  \mu^{(m)}_j}{\bar{\sigma}_j+ \epsilon_B}      
- \frac{  \mu^{(m)}_j}{\sigma^{(m)}_j+ \epsilon_B}
\right\rvert 
+
\sum_{m=M}^{\infty}
\dfrac{(\bar{\mu}+1)}{\epsilon_B^2}
\lvert \bar{\sigma}_j - \sigma^{(m)}_j\rvert,
\end{aligned}
\end{equation}
where the first term is finite by the fact that $ M $ is a finite constant. 
We have shown the condition for the second term to be finite in Lemma \ref{lem:sum_sigma_infty_m_finite}.
Therefore,
\begin{equation*}
\sum_{m=1}^{\infty}
\left\lvert
\frac{  \bar{\mu}_j}{\bar{\sigma}_j+ \epsilon_B}      
- \frac{  \mu^{(m)}_j}{\sigma^{(m)}_j+ \epsilon_B}
\right\rvert
< \infty , \forall j .
\end{equation*}

By \eqref{stp:prf2_finite_part2} and \eqref{stp:prf2_finite_part1}, we have that the right-hand side of \eqref{prf:2_4_6_true} is finite. It means that the left-hand side of \eqref{prf:2_4_6_true} is finite. Thus,

\hfill
$\displaystyle \sum_{m=1}^{\infty}
\left|
\bar{f}(\theta^{(m)},\lambda^{(m)}) - \bar{f}(\theta^{(m)}, \bar{\lambda})  
\right| < \infty. $ $ \qedwhite $

\begin{lemma}
	\label{lem:nonconvex_dimnishing2_last_lemma}
	If
	\[ \sum_{m=1}^{\infty} \sum_{i=m}^{\infty} \sum_{n=1}^{i} \alpha^{(i)} \eta^{(n)} <\infty \quad \text{and} \quad  
	\sum_{m=1}^{\infty} \sum_{n=m}^{\infty} \alpha^{(n)} <\infty , \]
	then
	\[  \limsup\limits_{M\rightarrow \infty}  \sum_{m=1}^{M} \eta^{(m)} \lVert \nabla \bar{f} (\theta^{(m)}, {\bar{\lambda}} ) \rVert_2^2    < \infty .\]
\end{lemma}
\textit{Proof.}
For simplicity of the proof, 
we define 
\[ T^{(M)} := \sum_{m=1}^{M} \eta^{(m)} \lVert \nabla \bar{f} (\theta^{(m)}, {\bar{\lambda}}) \rVert_2^2, \]
\[ O^{(m)} := \bar{f}(\theta^{(m+1)},\lambda^{(m+1)}) - \bar{f}(\theta^{(m)}, \lambda^{(m)}), \]
\[ \Delta_1^{(m+1)} := \bar{f}(\theta^{(m+1)},\lambda^{(m+1)}) - \bar{f}(\theta^{(m+1)}, \bar{\lambda}) , \]
\[ \Delta_2^{(m)} := \bar{f}(\theta^{(m+1)}, {\bar{\lambda}}) - \bar{f}(\theta^{(m)}, {\bar{\lambda}}), \]

where $ \bar{\lambda} $ is the converged value of $ \lambda $ in Theorem \ref{thm:lambda_converge}. Therefore, 

\begin{equation}
\label{prf:2_1}
O^{(m)} = \Delta_1^{(m+1)} +\Delta_1^{(m)} + \Delta_2^{(m)} \leq |\Delta_1^{(m+1)}| +|\Delta_1^{(m)}| + \Delta_2^{(m)}.
\end{equation}

By Proposition \ref{pro:lipschitz_2}, 
\begin{equation}
\Delta_2^{(m)} \leq - \eta^{(m)} \lVert \nabla \bar{f}(\theta^{(m)}, {\bar{\lambda}}) \rVert^2_2
+ \eta^{(m)} \bar{L} M \sqrt{n_2} a_m
+  \dfrac{1}{2} (\eta^{(m)})^2 \cdot   N\bar{L}M.
\label{prf:2_2}
\end{equation}

We sum the inequality \eqref{prf:2_1} from 1 to $ K $ with respect to $ m $ and plug \eqref{prf:2_2} into it to obtain
\begin{equation}
\begin{aligned}
\sum_{m=1}^{K} O^{(m)}
\leq& \sum_{m=1}^{K}  |\Delta_1^{(m+1)}|+ \sum_{m=1}^{K}  |\Delta_1^{(m)}| -\sum_{m=1}^{K} \{ \eta^{(m)} \lVert \nabla \bar{f} (\theta^{(m)}, {\bar{\lambda}}) \rVert^2_2 \} \\
&+ \sum_{m=1}^{K}\eta^{(m)} \bar{L} M \sqrt{n_2} a_m  + \sum_{m=1}^{K} \{\dfrac{1}{2} (\eta^{(m)})^2 N\bar{L}M\} \\ 
=& \sum_{m=1}^{K}  |\Delta_1^{(m+1)}|+ \sum_{m=1}^{K}  |\Delta_1^{(m)}| - T^{(K)} \\ &
+ \bar{L}^2 \sqrt{n_2} \cdot \sum_{m=1}^{K}\eta^{(m)}a_m +
\sum_{m=1}^{K} \{\dfrac{1}{2} (\eta^{(m)})^2 N\bar{L}M\} .
\end{aligned}
\label{prf:2_3}
\end{equation}
From this,  we have:
\begin{equation}
\begin{aligned}
\limsup\limits_{K\rightarrow \infty}
T^{(K)} 
\leq &  \limsup\limits_{K\rightarrow \infty} \dfrac{-1}{c_1}(\bar{f}(\theta^{(K)}, \lambda^{(K)} ) - \bar{f}(\theta^{(1)}, \lambda^{(1)})) 
\\  +&  \limsup\limits_{K\rightarrow \infty}  \dfrac{1}{c_1} \sum_{m=1}^{K}(|\Delta_1^{(m+1)}| +|\Delta_1^{(m)}|)  
\\  +&  \limsup\limits_{K\rightarrow \infty} \bar{L}^2 \sqrt{n_2}  \sum_{m=1}^{K}\eta^{(m)}a_m
\\  +&  \limsup\limits_{K\rightarrow \infty}
\dfrac{N\bar{L}K}{2c_1} \sum_{m=1}^{K} {\eta^{(m)}}^2.
\end{aligned}
\label{prf:2_3_T_decomposition}
\end{equation}

Next we show that each of the four terms in the right-hand side of \eqref{prf:2_3_T_decomposition} is finite, respectively.
For the first term, 
\begin{equation}
\limsup\limits_{K\rightarrow \infty}
\dfrac{-1}{c_1}(\bar{f}(\theta^{(K)}, \lambda^{(K)} ) - \bar{f}(\theta^{(1)}, \lambda^{(1)})) <\infty
\end{equation}

{ \cb is by the fact that the parameters $ \{\theta^{(m)}, \lambda^{(m)}\} $ are bounded by Assumption \ref{ass:bounded_para}, which implies that the image of 
$ f_i(\cdot) $ is in a bounded set.}

For the second term, we showed its finiteness in Lemma \ref{lemma:gap_bar_f_lambda_m_bar_finite}.

For the third term, by \eqref{dfn:a_m}, we have 
\begin{equation}
\begin{aligned}
& \limsup\limits_{K \rightarrow \infty} \sum_{m=1}^{K}  \eta^{(m)} a_m \\
= & \limsup\limits_{K \rightarrow \infty} \sum_{m=1}^{K}  \eta^{(m)} \left(
K_1
\sum_{i=m}^{\infty}
\sum_{j=1}^{i}
{\alpha}^{(i)} \eta^{(j)}
+ 
K_2
\sum_{i=m}^{\infty} \alpha^{(i)} 
\right) \\
=& K_1
\limsup\limits_{K \rightarrow \infty} 
\sum_{m=1}^{K} \eta^{(m)}
\left(
\sum_{i=m}^{\infty}
\sum_{j=1}^{i}
{\alpha}^{(i)} \eta^{(j)}
\right)
+ K_2
\limsup\limits_{K \rightarrow \infty} 
\sum_{m=1}^{K} \eta^{(m)}
\sum_{i=m}^{\infty} \alpha^{(i)} .
\end{aligned}
\label{prf:condition_to_get_rid_of_second_moment_assumption}
\end{equation}

The right-hand side of \eqref{prf:condition_to_get_rid_of_second_moment_assumption} is finite because 

\begin{equation}
\sum_{m=1}^{\infty} \eta^{(m)}
\left(
\sum_{i=m}^{\infty}
\sum_{j=1}^{i}
{\alpha}^{(i)} \eta^{(j)}
\right)
<
\sum_{m=1}^{\infty}
\left(
\sum_{i=m}^{\infty}
\sum_{j=1}^{i}
{\alpha}^{(i)} \eta^{(j)}
\right)
< \infty
\label{prf:dev_existed_assumption_1}
\end{equation}

and
\begin{equation}
\sum_{m=1}^{\infty} \eta^{(m)}
\sum_{i=m}^{\infty} \alpha^{(i)} 
<
\sum_{m=1}^{\infty}
\sum_{i=m}^{\infty} \alpha^{(i)} 
< \infty.
\label{prf:dev_existed_assumption_2}
\end{equation}

The second inequalities in \eqref{prf:dev_existed_assumption_1} and \eqref{prf:dev_existed_assumption_2} come from the stated assumptions of this lemma.

For the fourth term, 
\begin{equation}
\limsup\limits_{K\rightarrow \infty}
\dfrac{N\bar{L}M}{2c}\sum_{m=1}^{K} {\eta^{(m)}}^2 <\infty
\end{equation}
holds, because we have $ \sum_{m=1}^{\infty} (\eta^{(m)})^2 < \infty $ in Assumption \ref{ass:diminish_theta_update}.
Therefore, 
$  T^{(\infty)} = \sum_{m=1}^{\infty} \eta^{(m)} \lVert \nabla \bar{f} (\theta^{(m)}, {\bar{\lambda}}) \rVert_2^2 <\infty $ holds. \qedwhite

In Lemmas \ref{lem:sum_sigma_infty_m_finite},
\ref{lemma:gap_bar_f_lambda_m_bar_finite} and 
\ref{lem:nonconvex_dimnishing2_last_lemma}, we show that $ \{\sigma^{(m)}\} $ and $ \{\mu^{(m)} \} $ are Cauchy series, hence Lemma \ref{thm:nonconvex_diminishing2} holds.


%

\subsubsection*{Proof of Lemma \ref{thm:nonconvex_diminishing}}
This proof is similar to the the proof by \cite{Bertsekas}.

\textit{Proof.} By Theorem \ref{thm:nonconvex_diminishing2},  we have 
\begin{equation}
\limsup\limits_{M\rightarrow \infty}
\sum_{m=1}^{M} \eta^{(m)} \lVert \nabla \bar{f}(\theta^{(m)},\bar{\lambda}) \rVert_2^2 < \infty.
\label{prf:lemma_nonconvex_diminishing}
\end{equation} 
If there exists a $ \epsilon >0 $  and an integer $ \bar{m} $ such that 
\[ \lVert \nabla \bar{f}(\theta^{(m)},\bar{\lambda}) \rVert_2 \geq \epsilon  \] 
for all $ m \geq \bar{m} $, we would have 
\begin{equation}
\liminf\limits_{M\rightarrow \infty}
\sum_{m=\bar{m}}^{M} \eta^{(m)} \lVert \nabla \bar{f}(\theta^{(m)},\bar{\lambda}) \rVert_2^2
\geq
\liminf\limits_{M\rightarrow \infty}
\epsilon^2  \sum_{m=\bar{m}}^{M} \eta^{(m)} = \infty
\end{equation}
which contradicts \eqref{prf:lemma_nonconvex_diminishing}. 
Therefore, $ \liminf\limits_{m \rightarrow \infty} \lVert \nabla \bar{f}(\theta^{(m)},\bar{\lambda}) \rVert_2 = 0.$ $ \qedwhite $

\subsubsection*{Proof of Lemma \ref{lem:nonconvex_limit_gradient_zero}}

\begin{lemma} \label{lemma:berskas_lemma_1}
	Let $ Y_t, W,t $ and $ Z_t $ be three sequences such that $ W_t $ is nonnegative for all $ t $. Assume that 
	\begin{equation}
	Y_{t+1}\leq Y_t - W_t +Z_t, \quad t=0,1,...,
	\end{equation}
	and that the series $ \sum_{t=0}^{T} Z_t $ converges as $ T \rightarrow \infty $. Then either $ Y_t \rightarrow \infty $ or else $ Y_t $ converges to a finite value and $ \sum_{t=0}^{\infty} W_t <\infty $.
\end{lemma}

This lemma has been proven by \cite{Bertsekas}. 

\begin{lemma} \label{lemma:f_theta_m_bar_lambda_converge}
	When
	\begin{equation} \label{prf:condition:lemma:f_theta_m_bar_lambda_converge}
	\sum_{m=1}^{\infty} \sum_{i=m}^{\infty} \sum_{n=1}^{i} \alpha^{(i)} \eta^{(n)} <\infty \quad \text{and} \quad  
	\sum_{m=1}^{\infty} \sum_{n=m}^{\infty} \alpha^{(n)} <\infty ,
	\end{equation}
	it follows that $ \bar{f}(\theta^{(m)}, {\bar{\lambda}} ) $ converge to a finite value. 
\end{lemma}
\textit{Proof.}
By Proposition \ref{pro:lipschitz_2}, we have
\begin{equation}
\label{prf:stp:lemma:f_converge:ineq1}
\begin{aligned}
\bar{f}(\theta^{(m+1)}, \bar{\lambda} ) 
\leq & \bar{f}(\theta^{(m)} , \bar{\lambda} ) - \eta^{(m)} \lVert \nabla \bar{f}(\theta^{(m)},\bar{\lambda})   \rVert^2_2 \\
& + \eta^{(m)} \bar{L} M \sqrt{n_2} a_m
+  \dfrac{1}{2} (\eta^{(m)})^2 \cdot   N\bar{L}M.
\end{aligned}
\end{equation}

Let $ Y^{(m)} := \bar{f}(\theta^{(m)}, \bar{\lambda} ) $, $ W^{(m)}:=\eta^{(m)} \lVert \nabla \bar{f}(\theta^{(m)},\bar{\lambda})   \rVert^2_2 $ and $ Z^{(m)}:=  \eta^{(m)} \bar{L} M \sqrt{n_2} a_m
+  \dfrac{1}{2} (\eta^{(m)})^2 \cdot   N\bar{L}M $. 
By \eqref{res:diminish_theta_update} and \eqref{prf:condition_to_get_rid_of_second_moment_assumption}- \eqref{prf:dev_existed_assumption_2}, it is easy to see that $ \sum_{m=0}^{M} Z^{(m)} $ converges as $ M \rightarrow \infty $.
Therefore, by Lemma \ref{lemma:berskas_lemma_1}, $ Y^{(m)} $ converges to a finite value. The infinite case can not occur in our setting due to Assumptions  \ref{ass:lipschitz_theta} and  \ref{ass:bounded_para}.
$ \qedwhite $

\begin{lemma} \label{lemma:nonconex_limit_gradient_zero_last_lemma}
	If

$ 	\sum_{m=1}^{\infty} \sum_{i=m}^{\infty} \sum_{n=1}^{i} \alpha^{(i)} \eta^{(n)} <\infty \quad \text{and} \quad  
	\sum_{m=1}^{\infty} \sum_{n=m}^{\infty} \alpha^{(n)} <\infty , $

	then $ \lim\limits_{m \rightarrow \infty} \lVert \nabla \bar{f} (\theta^{(m)}, {\bar{\lambda}}) \rVert_2 = 0  $.
\end{lemma}
\textit{Proof.}
To show that $ \lim\limits_{m \rightarrow \infty} \lVert \nabla \bar{f} (\theta^{(m)}, {\bar{\lambda}}) \rVert_2 = 0  $, assume the contrary; that is, 
\[  \limsup\limits_{m \rightarrow \infty} \lVert \nabla \bar{f} (\theta^{(m)}, {\bar{\lambda}}) \rVert_2 > 0.   \]

Then there exists an $ \epsilon >0 $ such that $ \lVert \nabla \bar{f} (\theta^{(m)}, {\bar{\lambda}}) \rVert <\epsilon /2 $ for infinitely many $ m $ and also $ \lVert \nabla \bar{f} (\theta^{(m)}, {\bar{\lambda}}) \rVert >\epsilon $ for infinitely many $ m $. 
Therefore, there is an infinite subset of integers $ \mathbb{M} $, such that 
for each $ m \in \mathbb{M} $, there exists an integer $ q(m) > m  $ such that 
\begin{equation}
\begin{aligned}
\lVert \nabla \bar{f} (\theta^{(m)}, {\bar{\lambda}}) \rVert & < \epsilon/2, \\
\lVert \nabla \bar{f} (\theta^{(i(m))}, {\bar{\lambda}}) \rVert &> \epsilon,   \\
\epsilon/2 \leq \lVert \nabla \bar{f} (\theta^{(i)}, {\bar{\lambda}}) \rVert &\leq \epsilon, \quad  \\
\text{ if } m<i<q(m).&
\end{aligned}
\label{prf:lemma_413_condition}
\end{equation}
From
$ \begin{aligned}
\lVert \nabla \bar{f} (\theta^{(m+1)}, {\bar{\lambda}}) \rVert - \lVert \nabla \bar{f} (\theta^{(m)}, {\bar{\lambda}}) \rVert
\leq \bar{L} \eta^{(m)} \lVert \nabla \bar{f} (\theta^{(m)}, {{\lambda}^{(m)}})  \rVert,
\end{aligned} $
it follows that for all $ m\in \mathbb{M} $ that are sufficiently large so that $ \bar{L} \eta^{(m)} <\epsilon/4 $, we have 
\begin{equation}
\epsilon/4 \leq  \lVert \nabla \bar{f} (\theta^{(m)}, {{\lambda}^{(m)}})  \rVert .
\label{why_do_i_need_this_2}
\end{equation}

Otherwise the condition $ \epsilon/2 \leq \lVert \nabla \bar{f} (\theta^{(m+1)}, {\bar{\lambda}})  \rVert $ would be violated. Without loss of generality, we assume that the above relations as well as \eqref{prf:bound_diff_m+1_m_f} hold for all $ m \in \mathbb{M} $.
With the above observations, we have for all $ m \in \mathbb{M} $, 

\begin{equation}
\begin{aligned}
\dfrac{\epsilon}{2}
&\leq  \lVert \nabla \bar{f} (\theta^{q(m)}, {\bar{\lambda}}) \rVert - \lVert \nabla \bar{f} (\theta^{(m)}, {\bar{\lambda}}) \rVert \leq \bar{L} \lVert \theta^{q(m)}  - \theta^{(m)} \rVert \\
& \leq \bar{L} \sum_{i=m}^{q(m)-1} \eta^{(i)} (\lVert \nabla \bar{f} (\theta^{(i)}, \bar{\lambda})  \rVert + 
\lVert \nabla \bar{f} (\theta^{(i)}, {\lambda}^{(i)})  - \nabla \bar{f} (\theta^{(i)}, \bar{\lambda})\rVert) \\
& = \bar{L} \epsilon\sum_{i=m}^{q(m)-1} \eta^{(i)} + 
\bar{L}^2 \sqrt{n_2} M_1 \sum_{i=m}^{q(m)-1} \eta^{(i)} 
\sum_{j=m}^{\infty}  \sum_{k=1}^{j} {\alpha}^{(j)} \eta^{(k)}
\\
& + \bar{L}^2 \sqrt{n_2} M_2\sum_{i=m}^{q(m)-1} \eta^{(i)}
\sum_{j=m}^{\infty} \alpha^{(j)} \\
\end{aligned}
\end{equation}

The first inequality is by \eqref{prf:lemma_413_condition} and the third one is by the Lipschitz condition assumption. The seventh one is by \eqref{prf:pro:84:1}. 
By \eqref{condition:thm4.7}, we have for all $ m \in \mathbb{M}$, 

\begin{equation}
\sum_{i=m}^{q(m)-1} \eta^{(i)} 
\sum_{j=m}^{\infty}  \sum_{k=1}^{j} {\alpha}^{(j)} \eta^{(k)} 
<  \sum_{i=1}^{\infty} \sum_{j=i}^{\infty} \sum_{k=1}^{j} \alpha^{(j)} \eta^{(k)} <\infty
\end{equation}
and
\begin{equation}
\sum_{i=m}^{q(m)-1} \eta^{(i)}
\sum_{j=m}^{\infty} \alpha^{(j)} <
\sum_{i=1}^{\infty} \sum_{j=i}^{\infty} \alpha^{(j)} <\infty.
\end{equation}

It is easy to see that for any sequence $ \{ \alpha_i \} $ with $ \sum_{i=1}^{\infty} \alpha_i < \infty $, if follows that $ \liminf\limits_{M \rightarrow \infty} \sum_{i=M}^{\infty} \alpha_i = 0 $.  Therefore,
$ \liminf\limits_{m \rightarrow \infty}
\sum_{i=m}^{q(m)-1} \eta^{(i)} 
\sum_{j=m}^{\infty}  \sum_{k=1}^{j} {\alpha}^{(j)} \eta^{(k)}  = 0 $
and
$ \liminf\limits_{m \rightarrow \infty}
\sum_{i=m}^{q(m)-1} \eta^{(i)}
\sum_{j=m}^{\infty} \alpha^{(j)} =  0. $
From this it follows that
\begin{equation} \label{prf:wrong_statement}
\begin{aligned}
\liminf\limits_{m\rightarrow\infty} \sum_{i=m}^{q(m)-1} \eta^{(i)} 
\geq \dfrac{1}{2\bar{L}} .
\end{aligned}
\end{equation}

By \eqref{prf:pro:84:1} and  \eqref{why_do_i_need_this_2}, 
if we pick $ m\in \mathbb{M} $ such that $ L \sqrt{n_2}  a_m \leq  \dfrac{\epsilon}{8} $, we have 
$ \lVert \nabla \bar{f}(\theta^{(m)}, \bar{\lambda}) \rVert \geq \dfrac{\epsilon}{8}. $
Using \eqref{prf:bound_diff_m+1_m_f}, we observe that 
\begin{equation}
\begin{aligned}
& \bar{f}(\theta^{q(m)},{\bar{\lambda}} ) \\
& \leq 
\bar{f}(\theta^{(m)} , {\bar{\lambda}}) - c_1 \left(\dfrac{\epsilon}{8}\right)^2 \sum_{i=m}^{q(m)-1} \eta^{(i)}
+  \dfrac{1}{2}   \cdot N\bar{L}M  \sum_{i=m}^{q(m)-1} (\eta^{(i)})^2 , \forall m \in \mathbb{M},
\end{aligned}
\label{prf:413_final}
\end{equation}
where the second inequality is by \eqref{why_do_i_need_this_2}.
By Lemma \ref{lemma:f_theta_m_bar_lambda_converge}, $ \bar{f}(\theta^{q(m)},{\bar{\lambda}} )$ and 
$ \bar{f}(\theta^{(m)}, {\bar{\lambda}} ) $ converge to the same finite value. Using this convergence result and the assumption $ \sum_{m=0}^{\infty} (\eta^{(m)})^2 < \infty$, this relation implies that 

$  \limsup\limits_{m\rightarrow \infty,m\in \mathbb{M}}  \sum_{i=m}^{q(m)-1} \eta^{(i)} = 0  $
and contradicts \eqref{prf:wrong_statement}.   $ \qedwhite $

By Lemmas \ref{lemma:berskas_lemma_1},
\ref{lemma:f_theta_m_bar_lambda_converge} and 
\ref{lemma:nonconex_limit_gradient_zero_last_lemma}, we show that Theorem \ref{thm:double_limits} holds.

\subsection*{Discussions of conditions for stepsizes}

Here we discuss the actual conditions for $ \eta^{(m)} $ and $ \alpha^{(m)} $ to satisfy the assumptions of Theorem \ref{thm:lambda_converge} and Lemma \ref{thm:nonconvex_diminishing2}. We only consider the cases $ \eta^{(m)} = \frac{1}{m^k} $ and $ \alpha^{(m)} = \frac{1}{m^h} $, but the same analysis applies to the cases $ \eta^{(m)} = O(\frac{1}{m^k}) $ and $ \alpha^{(m)} = O(\frac{1}{m^h}) $.

\subsection*{Assumptions of Theorem \ref{thm:lambda_converge}}

For the assumptions of Theorem \ref{thm:lambda_converge},  the first condition
$   \sum_{m=1}^{\infty} \alpha^{(m)} <\infty  $
requires $ h>1 $.
Besides, the second condition 
\begin{equation}
\begin{aligned}
\sum_{m=1}^{\infty} \sum_{n=1}^{m} \alpha^{(m)} \eta^{(n)} 
\approx \dfrac{1}{h-1} \sum_{n=1}^{\infty} \eta^{(n)} \dfrac{1}{n^{h-1}} 
= \dfrac{1}{h-1} \sum_{n=1}^{\infty} \dfrac{1}{n^{k+h-1}} < \infty
\end{aligned}
\end{equation}
requires $ k+h>2 $.
The approximation comes from the fact that for every $ p>1, $ we have
\begin{equation}
\sum_{k=n}^{\infty} {k^{-p}} 
\approx \int_{k=n}^{\infty} {k^{-p}} dx 
= \left. \dfrac{1}{1-p} x^{1-p} \right\rvert^{\infty}_{n}
= \dfrac{1}{p-1} \dfrac{1}{n^{p-1}} .
\label{prf:approxiamte_inf_sum}
\end{equation}
Since $ k\geq 1 $ due to Assumption \ref{ass:diminish_theta_update}, we conclude that $ k+h>2.$
Therefore, the conditions for  $ \eta^{(m)} $ and $ \alpha^{(m)} $ to satisfy the assumptions of Theorem \ref{thm:lambda_converge} are $ h>1 $ and $ k\geq1 $. 

\subsection*{Assumptions of Lemma \ref{thm:nonconvex_diminishing2} }

For the assumptions of Theorem \ref{thm:lambda_converge},  the first condition
\begin{equation}
\sum_{m=1}^{\infty} \sum_{n=m}^{\infty} \alpha^{(n)} \approx   \sum_{m=1}^{\infty} \frac{1}{m^{h-1}}<\infty 
\label{prf:condition_lemma_nonconvex_stepsize}
\end{equation}

requires $ h>2 $.

Besides, the second condition is
\begin{equation}
\begin{aligned}
\sum_{m=1}^{\infty} \sum_{i=m}^{\infty} \sum_{n=1}^{i} \alpha^{(i)} \eta^{(n)}
=\sum_{m=1}^{\infty} \sum_{i=m}^{\infty} \alpha^{(i)} \sum_{n=1}^{i}  \eta^{(n)}
\leq C \sum_{m=1}^{\infty} \sum_{i=m}^{\infty}  \alpha^{(i)} <\infty.
\end{aligned}
\end{equation}
The inequality holds because for any $ p>1 $, we have
\begin{equation}
\sum_{k=1}^{n} k^{-p}
\approx \int_{k=1}^{n} k^{-p} dk
= \left. \dfrac{1}{1-p} k^{1-p} \right\rvert^{n}_{1}
= \dfrac{1}{p-1} (1-  n^{1-p}) 
\leq C
\label{prf:approxiamte_inf_sum_2}
\end{equation}

Therefore, the conditions for  $ \eta^{(m)} $ and $ \alpha^{(m)} $ to satisfy the assumptions of Lemma \ref{thm:nonconvex_diminishing2} are $ h>2 $ and $ k\geq1 $. 
}

\vskip 0.2in
\bibliography{library}

\begin{thebibliography}{21}
\providecommand{\natexlab}[1]{#1}
\providecommand{\url}[1]{\texttt{#1}}
\expandafter\ifx\csname urlstyle\endcsname\relax
  \providecommand{\doi}[1]{doi: #1}\else
  \providecommand{\doi}{doi: \begingroup \urlstyle{rm}\Url}\fi

\bibitem[kdd(1999)]{kdd99}
{KDD Cup 1999 Data}, 1999.
\newblock URL \url{http://www.kdd.org/kdd-cup/view/kdd-cup-1999/Data}.

\bibitem[Arpit et~al.(2016)Arpit, Zhou, Kota, and Govindaraju]{Arpit2016}
Devansh Arpit, Yingbo Zhou, Bhargava~U. Kota, and Venu Govindaraju.
\newblock {Normalization Propagation: A Parametric Technique for Removing
  Internal Covariate Shift in Deep Networks}.
\newblock In \emph{International Conference on Machine Learning}, volume~48,
  page~11, 2016.

\bibitem[Ba et~al.(2016)Ba, Kiros, and Hinton]{Ba}
Jimmy~Lei Ba, Jamie~Ryan Kiros, and Geoffrey~E. Hinton.
\newblock {Layer Normalization}.
\newblock \emph{arXiv preprint arXiv:1607.06450}, 2016.

\bibitem[Bertsekas(2011)]{Bertsekas2010}
Dimitri~P. Bertsekas.
\newblock {Incremental gradient, subgradient, and proximal methods for convex
  optimization: A Survey}.
\newblock \emph{Optimization for Machine Learning}, 2010\penalty0 (3):\penalty0
  1--38, 2011.

\bibitem[Bertsekas and Tsitsiklis(2000)]{Bertsekas}
Dimitri~P. Bertsekas and John~N. Tsitsiklis.
\newblock {Gradient Convergence in Gradient Methods with Errors}.
\newblock \emph{SIAM Journal on Optimization}, 10:\penalty0 627--642, 2000.

\bibitem[Bjorck et~al.(2018)Bjorck, Gomes, Selman, and
  Q.~Weinberger]{Johan2018}
Johan Bjorck, Carla Gomes, Bart Selman, and Kilian Q.~Weinberger.
\newblock Understanding batch normalization.
\newblock \emph{arXiv preprint arXiv:1806.02375}, 2018.

\bibitem[Bottou et~al.(2016)Bottou, Curtis, and Nocedal]{Bottou2016}
L{\'{e}}on Bottou, Frank~E. Curtis, and Jorge Nocedal.
\newblock {Optimization Methods for Large-Scale Machine Learning}.
\newblock \emph{arXiv preprint arXiv:1606.04838}, 2016.

\bibitem[Cooijmans et~al.(2016)Cooijmans, Ballas, Laurent, and
  Courville]{Cooijmans}
Tim Cooijmans, Nicolas Ballas, C{\'{e}}sar Laurent, and Aaron Courville.
\newblock {Recurrent Batch Normalization}.
\newblock \emph{arXiv preprint arXiv:1603.09025}, 2016.

\bibitem[{Duchi, John and Hazan, Elad and Singer}(2011)]{Duchi}
Yoram {Duchi, John and Hazan, Elad and Singer}.
\newblock {Adaptive Subgradient Methods for Online Learning and Stochastic
  Optimization}.
\newblock \emph{Journal of Machine Learning Research}, 12\penalty0
  (Jul):\penalty0 2121--2159, 2011.

\bibitem[Grippo(1994)]{Grippo1994}
L.~Grippo.
\newblock {A Class of Unconstrained Minimization Methods for Neural Network
  Training}.
\newblock \emph{Optimization Methods and Software}, 4\penalty0 (2):\penalty0
  135--150, 1994.

\bibitem[G{\"{u}}l{\c{c}}ehre and Bengio(2016)]{Gulcehre2016}
{\c{C}}aglar G{\"{u}}l{\c{c}}ehre and Yoshua Bengio.
\newblock {Knowledge Matters: Importance of Prior Information for
  Optimization}.
\newblock \emph{Journal of Machine Learning Research}, 17\penalty0
  (8):\penalty0 1--32, 2016.

\bibitem[He et~al.(2016)He, Zhang, Ren, and Sun]{He2015a}
Kaiming He, Xiangyu Zhang, Shaoqing Ren, and Jian Sun.
\newblock {Deep Residual Learning for Image Recognition}.
\newblock In \emph{Computer Vision and Pattern Recognition}, pages 770--778,
  dec 2016.

\bibitem[Ioffe and Szegedy(2015)]{SergeyIoffe2015}
Sergey Ioffe and Christian Szegedy.
\newblock {Batch Normalization: Accelerating Deep Network Training by Reducing
  Internal Covariate Shift}.
\newblock In \emph{International Conference on Machine Learning}, pages
  448--456, 2015.

\bibitem[Krizhevsky and Hinton(2009)]{Krizhevsky2009}
Alex Krizhevsky and Geoffrey~E. Hinton.
\newblock \emph{{Learning Multiple Layers of Features from Tiny Images}}.
\newblock PhD thesis, 2009.

\bibitem[Krizhevsky et~al.(2012)Krizhevsky, Sutskever, and
  Hinton]{Krizhevsky2012}
Alex Krizhevsky, Ilya Sutskever, and Geoffrey~E. Hinton.
\newblock {Imagenet Classification with Deep Convolutional Neural Networks}.
\newblock In \emph{Advances in neural information processing systems}, pages
  1097--1105, 2012.

\bibitem[LeCun et~al.(1998)LeCun, Bottou, and Bengio]{LeCun1998b}
Yann LeCun, L{\'{e}}on Bottou, and Yoshua Bengio.
\newblock {Gradient-based learning applied to document recognition}.
\newblock \emph{Proceedings of the IEEE}, 86\penalty0 (11):\penalty0
  2278--2324, 1998.

\bibitem[Orr and M{\"{u}}ller(2003)]{Orr2003}
Genevieve~B. Orr and Klaus-Robert M{\"{u}}ller.
\newblock \emph{{Neural Networks: Tricks of the Trade}}.
\newblock Springer, New York, 2003.

\bibitem[Polyak(1987)]{Polyak1987}
B.~T. Polyak.
\newblock {Introduction to optimization. Translations series in mathematics and
  engineering}.
\newblock \emph{Optimization Software}, 1987.

\bibitem[Polyak and Tsypkin(1973)]{Polyak1973}
B.~T. Polyak and Y.~Z. Tsypkin.
\newblock {Pseudogradient Adaption and Training Algorithms}.
\newblock \emph{Automation and Remote Control}, 34:\penalty0 45--67, 1973.

\bibitem[Salimans and Kingma(2016)]{Salimans}
Tim Salimans and Diederik~P. Kingma.
\newblock {Weight Normalization: A Simple Reparameterization to Accelerate
  Training of Deep Neural Networks}.
\newblock In \emph{Advances in Neural Information Processing Systems}, pages
  901--901, 2016.

\bibitem[Santurkar et~al.(2018)Santurkar, Tsipras, Ilyas, and
  Madry]{santurkar2018does}
Shibani Santurkar, Dimitris Tsipras, Andrew Ilyas, and Aleksander Madry.
\newblock How does batch normalization help optimization?
\newblock \emph{arXiv preprint arXiv:1805.11604}, 2018.

\end{thebibliography}

\end{document}